\documentclass{article}
\usepackage[table,dvipsnames]{xcolor}
\usepackage[utf8]{inputenc}
\usepackage{amsmath}
\usepackage{tikz}
\usepackage{tkz-euclide}
\usepackage[margin=0.75in]{geometry}
\usepackage [english]{babel}
\usepackage [autostyle, english = american]{csquotes}
\usepackage{caption}
\usepackage{subcaption}
\MakeOuterQuote{"}
\usepackage{wasysym}
\usepackage{multirow}
\usepackage{amssymb}

\numberwithin{figure}{section}

\title{A Survey of Inductive Biases for Factorial Representation-Learning}
\author{Karl Ridgeway}
\date{November 2016}

\usepackage{graphicx}

\usepackage{titlesec}
\titleclass{\subsubsubsection}{straight}[\subsection]
\newcounter{subsubsubsection}[subsubsection]
\renewcommand\thesubsubsubsection{\thesubsubsection.\arabic{subsubsubsection}}
\titleformat{\subsubsubsection}
  {\normalfont\normalsize\bfseries}{\thesubsubsubsection}{1em}{}
\titlespacing*{\subsubsubsection}
{0pt}{3.25ex plus 1ex minus .2ex}{1.5ex plus .2ex}
\makeatletter
\renewcommand\paragraph{\@startsection{paragraph}{5}{\z@}%
  {3.25ex \@plus1ex \@minus.2ex}%
  {-1em}%
  {\normalfont\normalsize\bfseries}}
\renewcommand\subparagraph{\@startsection{subparagraph}{6}{\parindent}%
  {3.25ex \@plus1ex \@minus .2ex}%
  {-1em}%
  {\normalfont\normalsize\bfseries}}
\def\toclevel@subsubsubsection{4}
\def\toclevel@paragraph{5}
\def\toclevel@paragraph{6}
\def\l@subsubsubsection{\@dottedtocline{4}{7em}{4em}}
\def\l@paragraph{\@dottedtocline{5}{10em}{5em}}
\def\l@subparagraph{\@dottedtocline{6}{14em}{6em}}
\makeatother
\usepackage{etoolbox}
\newtoggle{ica}
\newtoggle{nmf}
\newtoggle{pca}
\newtoggle{isa}
\newtoggle{cvq}
\newtoggle{multilinear}
\newtoggle{rica}
\newtoggle{vae}
\newtoggle{sparse_autoencoder}
\newtoggle{autoencoder}
\newtoggle{dva}
\newtoggle{karaletsos}
\newtoggle{mm_t_sne}
\newtoggle{tae}
\newtoggle{siamese}
\newtoggle{dc_ign}
\newtoggle{ssvae}
\newtoggle{disbm}

\toggletrue{ica}
\toggletrue{nmf}
\toggletrue{pca}
\toggletrue{isa}
\toggletrue{cvq}
\toggletrue{multilinear}
\toggletrue{rica}
\toggletrue{vae}
\toggletrue{sparse_autoencoder}
\toggletrue{autoencoder}
\toggletrue{dva}
\toggletrue{karaletsos}
\toggletrue{mm_t_sne}
\toggletrue{tae}
\toggletrue{siamese}
\toggletrue{dc_ign}
\toggletrue{ssvae}
\toggletrue{disbm}

\begin{document}
\begin{abstract}
With the resurgence of interest in neural networks, representation learning has
re-emerged as a central focus in artificial intelligence.  Representation
learning refers to the discovery of useful encodings of data that make
domain-relevant information explicit.  \emph{Factorial representations}
identify underlying independent causal factors of variation in 
data. A factorial representation is compact and faithful, makes the causal
factors explicit, and facilitates human interpretation of data.  Factorial
representations support a variety of applications, including the generation
of novel examples, indexing and search, novelty detection, and transfer 
learning.

This article surveys various constraints that encourage a
learning algorithm to discover factorial representations. I dichotomize
the constraints in terms of unsupervised and supervised inductive bias.
Unsupervised inductive biases exploit assumptions about the environment,
such as the statistical distribution of factor coefficients, assumptions
about the perturbations a factor should be invariant to (e.g. a representation
of an object can be invariant to rotation, translation or scaling),
and assumptions about how factors are combined to synthesize an observation.
Supervised inductive biases are constraints on the representations based on
additional information connected to observations. Supervisory labels come
in variety of types, which vary in how strongly they constrain the
representation, how many factors are labeled, how many observations are
labeled, and whether or not we know the associations between the constraints
and the factors they are related to.

This survey brings together a wide variety of models that all touch on the
problem of learning factorial representations and lays
out a framework for comparing these models based on the strengths of the
underlying supervised and unsupervised inductive biases.  
\end{abstract}

\maketitle
\tableofcontents

\newcommand{\gen}[1]{\noindent \textcolor{ForestGreen}{\textbf{#1}}}
\newcommand{\ind}[1]{\noindent \textcolor{Blue}{\textbf{#1}}}
\newcommand{\rep}[1]{\noindent \textcolor{Sepia}{\textbf{#1}}}
\newcommand{\examp}[1]{\noindent \textcolor{Bittersweet}{\textbf{#1}}}
\newcommand{\othermodels}[1]{\noindent \textcolor{RedViolet}{\textbf{#1}}}

\section{Introduction}
With the advent of modern deep neural nets, representation learning has re-emerged as a primary focus
in machine learning. Representation learning refers to the discovery of useful representations of data.
Data samples are drawn from some environment --- for example, photographs of faces or audio recordings of music. 
A data representation is a means for transforming these raw examples into some new space. Typically, we have some particular task in mind,
and the data representation allows us to easily recover attributes from the data relevant to the task at hand.
Suppose the task is to create a searchable index of the dataset. The best representation for this task might
summarize the face photographs as a set of attributes --- ethnicity, hair color, or eye color.  We might similarly
represent music by its tempo, genre, artist, or key signature.
A good representation makes explicit relevant information about the task. While ethnicity and hair color 
are, in some way, represented in the raw pixels of the image, as is tempo in the audio recording, the more 
explicit representation makes acting on the information much easier.

\subsection{Desiderata for Representations}
Different tasks and environments place different demands on representations, but there are a few characteristics
commonly associated with good representations which are useful across a wide variety of tasks.
Good representations are both \emph{compact} and \emph{faithful} to information represented in the input.
They also \emph{explicitly} represent the attributes required for the task at hand.
Finally, they are \emph{interpretable} by humans.

\subsubsection{Compact and Faithful}
A faithful representation preserves the information in the observation \cite{smolensky1990tensor},
with as little distortion as possible.
A perfectly faithful representation can of course be achieved
by not doing any transformation on the input --- this is why good representations also need to encode 
the information compactly. This is equivalent to the representation being an efficient compression
of the input.

The faithfulness of a representation can only be evaluated in terms of the task it's intended to be used
with. Different tasks have different demands on how much of the input need be represented. A classification
task might just need to know object category. An image compression program should represent
all of the non-noise features in the image. A representation of medical test results should probably be
fully invertible, meaning we can perfectly reconstruct the input from the representation.

\subsubsection{Explicit}
\begin{figure}[ht]
\center
\includegraphics[scale=0.15]{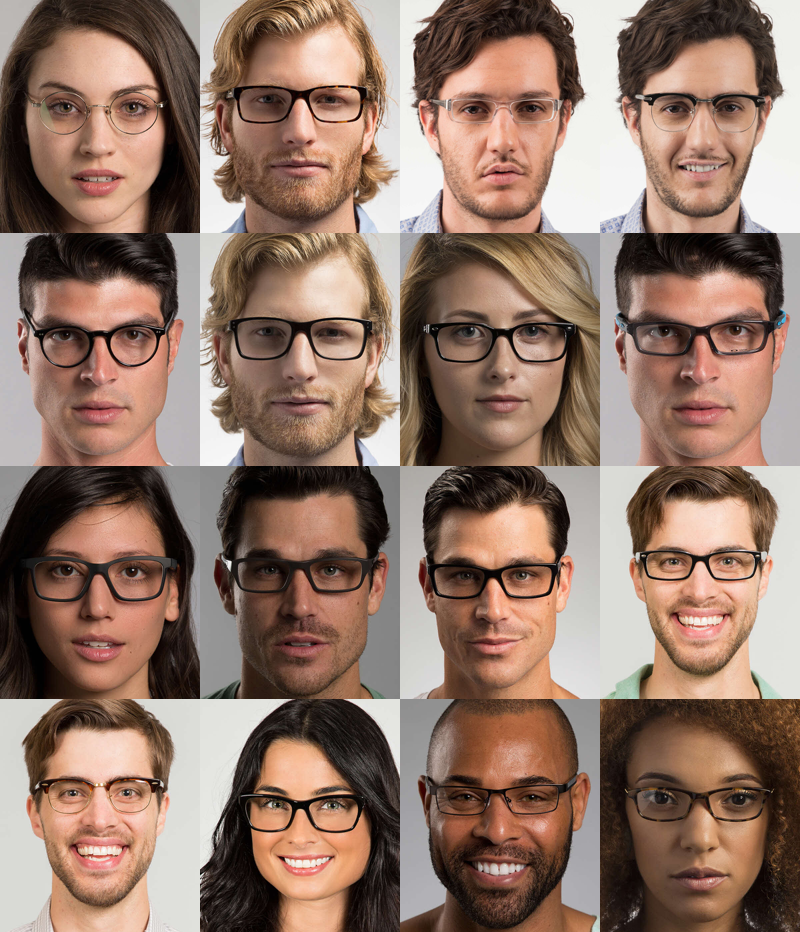}
\caption{Random draws from the dataset of subjects wearing various styles of glasses.}
\label{fig:framesdirect}
\end{figure}

The representation should explicitly encode the attributes necessary for whatever task is at hand,
without interference from other attributes or sources of noise. Attributes can be
far removed from the low-level observation;
for example, facial expression is observed as a complex conjunction of muscle movements around 
the mouth, the eyes, and jaw. Attributes are often combined in complex ways with other attributes.
Consider how the glasses in Figure~\ref{fig:framesdirect} interact 
with identity, pose, and lighting conditions --- they fit onto and cast a shadow on the face.
Explicit representations of attributes should be \emph{invariant} to these kinds of variations ---
identity should not change if the image is slightly corrupted by noise or if the light source is 
moved, even  though many pixels on the face change.
Explicit representations are also invariant to small local changes / noise in the input.

A representation of an attribute that makes some information explicit should have a \emph{structure} that reflects
the information represented:
A categorical, discrete, or continuous attribute needs to be modeled
as a variable of the appropriate type. Additionally, attributes can be single- or multi-dimensional.
For example, color can be represented in a single dimension (frequency) or as a three-dimensional 
vector (red/green/blue channels). The choice of variable type depends on the environment and task.

\subsubsection{Interpretable}
Many tasks rely on data representations being \emph{interpretable} by humans. 
Representations are often better off if they can be easily 
interpreted by humans, regardless of whether humans will use them directly or not. For example,
the task of predicting face gender will be much easier if the presence of facial hair and makeup
is directly represented in the code.

Representations that are interpretable need to be unique. A representation that can be exchanged 
for another that is equally valid is more difficult to attach meaning to. They should also be 
identifiable, meaning that they are tied to some named concept or attribute.

\

I have described three desirable properties of representations --- compactness/faithfulness, expressivity, and intepretability.
Now, I will describe a type of representation, called a factorial representation, that can help achieve these goals.

\subsection{Factorial Representations}

In a \emph{factorial representation}, the attributes are statistically independent. 
A \emph{factor} is one of these independent attributes.
Imagine a set of images of landscapes containing trees and sheep. In the generative
story of these landscapes, the location and number of sheep is largely independent of the location and number of
trees. Another common example is that of blind source separation --- in a room full of people speaking, we can
assume that the probability of any one person speaking is independent of others speaking simultaneously.
This independence assumption places restrictions on the environment --- that the factors are truly independent ---
and on the task, which should depend on these independent causes. For the right environment and task,
factorial representations provide numerous benefits: they provide a useful bias for 
learning \cite{zemel1993minimum} and provide more compact/faithful, explicit, and interpretable representations.
Horace Barlow argued for the biological and computational utility of factorial representations 
\cite{barlow1961,Barlow1989UnsupervisedLearning}.

\subsubsection{Statistical Independence}
\begin{figure}[ht]
\center
\includegraphics[scale=0.6]{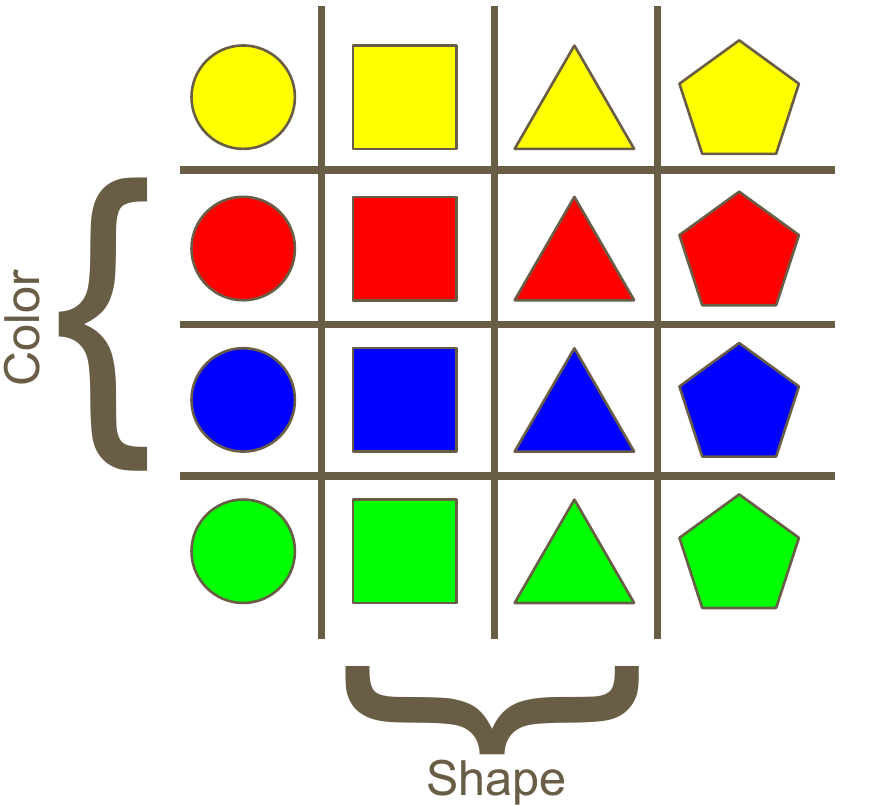}
\caption{An example dataset with two independent factors of variation: shape and color.}
\label{fig:shapes_and_colors}
\end{figure}

When variables are statistically independent, the joint probability distribution is factorial. 
Consider the example set of colored shapes in Figure~\ref{fig:shapes_and_colors}. Each
observation can be encoded as a combination of a shape and color. 
The occurrence probability of any shape or color is 
one-fourth, and the occurrence probability of any combination of the two, one-sixteenth.
This two-factor representation is \emph{factorial} because the joint distribution $P(shape,color)$ can be
factorized into the product of the distributions of the components $P(shape) P(color)$ --- consequently, the
component factors are statistically independent of one another, and have no redundancy.

\subsubsection{Useful Properties of Factorial Representations}
A factorial representation that is faithful to the factors modeled minimizes redundancy between them.
Therefore, these representations are also maximally compact. 
Factorial representations are also more interpretable: if the representation explicitly captures the true
independent generative factors, these factors should be unique and identifiable. 
The shape and color example illustrates this nicely --- there is no other way to describe these images that
is as compact, faithful, explicit, and interpretable as "shape and color". Independent factors need to be 
invariant with respect to one another --- no change in shape should alter our impression of an image's color.

\noindent These properties make factorial representations very useful for certain types of tasks, such as:
\begin{itemize}
\item \textbf{Novel example generation}. An invertible factorial representation can be used to generate
novel examples not found in the original dataset. In fact, we can manipulate
the representation to generate examples of the full Cartesian product of 
factor combinations --- we could generate men with the women's styles of glasses in Figure~\ref{fig:framesdirect}.
\item \textbf{Novelty Detection}. Since factors are independent, any statistical dependence found between 
factors in a  new dataset indicates a new situation \cite{Barlow1989}.
\item \textbf{Search}. A factorial representation also supports efficient searching 
--- in the  example shown in Figure~\ref{fig:shapes_and_colors},
specifying a shape narrows a search down to one-quarter of the dataset. 
\item \textbf{Prediction}. More generally, they are useful as features 
used as input to supervised learning methods, 
as observed by \cite{becker1991unsupervised} for linear supervised tasks.
\item \textbf{Compression}. Factorial representations are compact: a representation that 
minimizes redundancy also minimizes the cost associated with storing the representation, 
according to the minimum description length principle \cite{zemel1993minimum}.
\end{itemize}


\subsubsection{Factorial Representations in Non-Factorial Datasets and Environments}
The independence assumption of a factorial representation is easily violated. It can
be violated by the method used to sample from the environment, or by the environment itself.
Consider the case of missing data: If the dataset in Figure~\ref{fig:shapes_and_colors} 
were missing just one of the 16 shape/color combinations, then $P(shape,color) \neq P(shape)P(color)$. 
More generally, datasets can have a \emph{sampling bias}, in which certain combinations
of factors are systematically missing, or sampled more than other combinations. 
This imbalance in factors can of course also be caused by noise. The bias
imposed by a factorial representation can help to learn better representations in
the case of datasets with a sampling bias or noise.

Often, the environment itself is non-factorial. Datasets have natural dependencies --- 
for example, facial hair and gender are naturally dependent. A factorial representation
that teases apart facial hair and gender will over-generate: it will assume that facial 
hair and gender are independent, so a woman with a beard will be just as likely as a man 
with a beard, according to the representation. 

For some tasks, representing these
dependencies is critical, so a factorial representation is inappropriate. 
For example, a person's identity might be naturally correlated with the style
of glasses worn by that person. If our task is to generate example "typical" images of
people (what we might expect them to look like on any given day), we want to consider 
a person's glasses style as part of their identity.
For other tasks, we might want to ignore this dependency. If our task is high-security face verification,
we want to be sure that our program will not falsely accept an impostor
who wears the same glasses as the user. In this setting, imposing the bias of a factorial
representation is helpful.

\section{Inductive Biases}
In the last section, we noted that imposing the bias of a factorial representation can result 
in improved model explicitness as well as better interpretability. 
This paper explores different \emph{inductive biases} that can be used
to learn factorial representations for a variety of environments and tasks.

\begin{figure}[h!]
\center
\begin{tikzpicture}
    \draw [thin, gray, ->] (0,-0.1) -- (0,6);      
    \node [left,text width=6cm,text centered, font=\bf,rotate=90] at (-1.2,6.1) {Supervised Signal Strength};
    \node [left] at (0,0.5) {none};
    \node [left] at (0,1.5) {low};
    \node [left] at (0,5.5) {strong};
    
    \draw [thin, gray, ->] (-.1,0) -- (11,0);      
    \node [below, text centered, font=\bf] at (5,-0.5) {Unsupervised Bias};
    \node [below] at (0.75,-0.1) {none};
    \node [below] at (1.75,0) {low};
    \node [below] at (10.5,0) {high};
    
    
    \node [right,text width=2cm] at (1.2,0.5) {\textsc{Auto-encoder}};
    \node [right,text width=2cm] at (2.8,0.7) {\textsc{VAE, Sparse-\newline Autoencoder}};
    \node [right,text width=2.2cm] at (5.7,0.5) {\textsc{R-ICA, Multilinear}};
    \node [right,text width=1.2cm] at (8,0.5) {\textsc{ISA, CVQ}};
    \node [right] at (9,0.5) {\textsc{PCA}};
    \node [right,text width=1.2cm] at (10,0.5) {\textsc{ICA}, NMF};
    
    \node [right] at (2.8,5.5) {\textsc{SSVAE}};
    \node [right] at (5.7,5) {\textsc{disBM (1)}};
    \node [right, text width=1.8cm] at (2.8,4.5) {\textsc{DC-IGN}};
    \node [right, text width=1.8cm] at (1.2,4.5) {\textsc{Siamese}};
    \node [right] at (5.7,4.5) {\textsc{disBM (2)}};
    \node [right] at (1.2,3.5) {\textsc{TAE}};
    \node [right, text width=1.2cm] at (0.0,3.5) {\textsc{mm-\newline t-SNE}};
    \node [right] at (2.8,2.5) {\textsc{Karaletsos}};
    \node [right] at (1.2,1.5) {\textsc{DVA}};
\end{tikzpicture}
\caption{Model organization chart}
\label{fig:model_chart}
\end{figure}
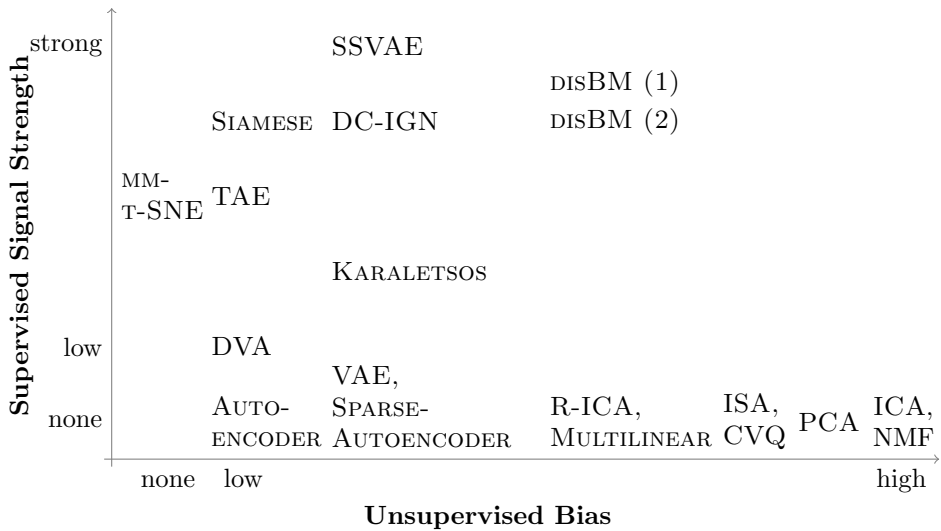

In this section, I will highlight several different approaches for biasing
representations to be more factorial. Each of these biases makes a different
assumption about the generative environment.
First, I will look at biases that
assume different \emph{distributions} of coefficients for a factor.
Next, I will look at biases that allow a factor to be \emph{invariant} to
changes in the output that are not related to the factor.
Another class of bias look at the ways that factors can \emph{combine} to
generate the observed data.
Finally, I will look at how to apply \emph{supervision} as a bias to
improve factorial representations.
Since supervision is mostly orthogonal to, and can be arbitrarily combined with 
the other types of bias, I have
given it its own dimension in the model chart in Figure~\ref{fig:model_chart}.
Unsupervised biases leverage assumptions about the generative environment to learn
factorial representations, while supervised biases are specific to a type of prior domain
knowledge.

Here, I focus on inductive biases that are highly generic, meaning that
they could apply to almost any observation domain, such as images, audio, 
or medical test results. 
The biases considered here allow us to learn
representations that are more explicit, and more interpretable --- 
these biases allow us to gain a better understanding of the underlying causal structure,
regardless of the observation domain. I will therefore avoid describing inductive biases
that are too specific to apply to a range of domains.

For consistency, I will introduce each unsupervised inductive bias the same way.
First, I will describe the \gen{generative assumptions} of the model. Then, I will
describe an \ind{inductive bias} that is consistent with this environment, and the corresponding
learned \rep{representation structure}. Finally, I will show some \examp{examples} of the 
inductive bias in action, and mention other models that use a similar inductive bias.
When possible, I will also show examples that incorporate multiple inductive biases simultaneously.

By matching the inductive biases of a model with the characteristics of the environment and
the requirements of the task, it is possible to discover factorial representations in many settings.

\subsection{Distributional Bias}
In this section, I will explore some inductive biases for discovering 
factorial representations that are motivated by the statistical distribution of the 
factors in the environment. To highlight the differences between these
distribution biases, I will mostly focus on methods that also use
a \emph{linear} bias --- an assumption that
factors are linearly related to, and are linearly combined in, the observation. 
Keeping this bias constant makes it easier to highlight the differences in distributional
bias.
Where applicable, I will also highlight non-linear models that incorporate these 
distributional biases. 

I will describe two different assumptions about the distribution of coefficients for factors
that each lead to a type of inductive bias. 
In some environments, factor coefficients follow a \emph{Gaussian} distribution.
I will describe an inductive bias that finds factorial representations when
the environment has Gaussian-distributed factors. 
I will show how naturalistic environments are often non-Gaussian, in that they are
sparse. Likewise, I will describe a bias that corresponds to sparse environments.
In sparse environments, I show that the sparsity bias produces representations that are 
better aligned to the true causal structure of the environment than representations 
with Gaussian-distributed factors.

\subsubsection{Gaussian Distribution of Factors}
\gen{Generative Assumptions.}\quad 
A factorial representation should ideally have non-redundant, statistically independent factors. 
Since two factors that are jointly Gaussian-distributed are independent when they are uncorrelated, one 
way to find independent factors in a Gaussian environment
is to reduce or minimize the correlation between all pairs of factors.
Principle Components Analysis, or \textsc{PCA}, is a model that learns representations 
with zero correlation between the latent factors. 

\subsubsubsection{PCA}
\ind{Inductive Bias.}\quad
\textsc{PCA}'s mapping function is linear: a 
($1 \times D$)-dimensional input vector $\mathrm{x}$ is transformed into its ($1 \times N$)-dimensional 
latent representation $\mathrm{z}$ by multiplying $x$ with a  $(D\times N)$ matrix $\mathrm{W}$: 
\begin{equation}\label{eq:pca}
    \mathbf{z} = \mathbf{xW}
\end{equation}
We can also compute the \emph{reconstruction}, or the prediction of $\mathrm{x}$ given $\mathrm{z}$.
\begin{equation}
    \mathbf{\tilde{x}} = \mathbf{zW^{-1}}
\end{equation}
\textsc{PCA} minimizes \emph{reconstruction cost}, or the euclidean distance between the input and the reconstruction:
\begin{equation}\label{eq:pcaloss}
    \mathcal{L}_{\mathrm{PCA}} = \lVert \mathbf{x - \tilde{x}} \rVert_2^2
\end{equation}
\textsc{PCA} also restricts the the dimensions of \textbf{z} to be uncorrelated --- the columns of \textbf{W} are all orthogonal.
Two variables that are jointly Gaussian are independent if they are uncorrelated. 
Therefore, decorrelation finds independent components in the case of Gaussian-distributed causes.
\textsc{PCA} is further constrained to consider only components that are linearly related to the input.\footnote{A family 
    of related models share similar characteristics --- probabilistic \textsc{PCA} adds isotropic
    Gaussian observation noise to equation \ref{eq:pca}. Factor analysis extends this further by adding
    diagonal instead of isotropic noise. Linear autoencoders are essentially a variant of \textsc{PCA} trained 
    with stochastic gradient descent.} 
\textsc{PCA} produces a list of components, ordered by the amount of variance in the input that they explain.
By considering only the top $M$ components, one can trade off between compactness and faithfulness of the
representation.

\rep{Representation Structure.}\quad \textsc{PCA}'s representation is very simple --- each observation
is mapped to a vector of scalar factor coefficients \textbf{z}.

Unfortunately, representations learned by \textsc{PCA} are not unique, in that representations are unique only to
rotations in the latent
space \cite{bishop2006pattern}. This means that the representation learned is often just one sample of many
possible representations that are equally valid according to \textsc{PCA}. This presents some obvious problems
when we try to interpret the results of \textsc{PCA}, since one solution can be exchanged for other, equally 
valid solutions. 

\subsubsubsection{Non-PCA}
There are other approaches for
reducing or removing correlations in representations. For example,
\cite{Cheung2014DiscoveringNetworks} added a "cross-covariance penalty" to the loss
function of an autoencoder. This extra penalty term seeks to directly minimize
the pairwise correlations between hidden units. In \cite{Foldiak1990FormingLearning}, the
author used direct inhibitory ("anti-Hebbian") feedback connections between the representation factors,
which has the effect of of decreasing correlations between factors.

A variational auto-encoder (\textsc{VAE}) \cite{Kingma2013Auto-EncodingBayes} adds a the bias 
of Gaussian-distributed factors to an autoencoder by imposing a Gaussian prior on the 
representation, in which the covariance matrix is (typically) diagonal. This corresponds to assuming
that the factors are Gaussian and uncorrelated/independent.
Autoencoders are nonlinear models and thus do not share the linear bias of \textsc{PCA},
so the \textsc{VAE} appears to the left of \textsc{PCA} in the model chart in Figure~\ref{fig:model_chart}. 

\subsubsection{Non-Gaussian Distribution of Factors}
\gen{Generative Assumptions.}\quad 
The factors in many environments are non-Gaussian. In this section, I will show
methods for finding factorial representations in non-Gaussian settings that also solve the 
uniqueness problem of \textsc{PCA}. First I will describe a theoretical motivation for why
decorrelation is insufficient in non-Gaussian settings, which leads to a generic method for finding
factorial representations called Independent Components Analysis or \textsc{ICA}. Then, I will describe 
a subset of non-Gaussian distributions, called sparse distributions, that are appropriate
for most naturalistic environments. These distributions lead to a specific type of \textsc{ICA} called \textsc{Sparse ICA}. 
I will then describe some Non-ICA methods for discovering non-Gaussian independent factors.

To find independent factors that are not Gaussian-distributed, it is essential to impose a stronger
bias than decorrelation.
Correlation between two random variables $\rho(x,y)$ depends only on the first moment of the joint PDF
$p(x,y)$. However, independence is a generalized measure of non-correlation: two variables are independent
if and only if $\rho(x^p,y^q) = 0$ for all positive integer values $p$ and $q$. For jointly Gaussian-distributed
variables, there is no higher-order correlation structure, and so non-correlation is a sufficient condition 
for independence.
However, the factors of variation in many naturalistic environments are non-Gaussian. Finding the true 
independent causes in these environments requires a stronger inductive bias.

\subsubsubsection{ICA}
\ind{Inductive Bias.}\quad 
An approach to finding truly independent factors is to find the most non-Gaussian factors that are also faithful
to the input data. Intuitively, if a learned component is actually the sum of more than one of the true 
independent, non-Gaussian factors, that sum will look more Gaussian, according to the central limit theorem. 
Only the true non-Gaussian source factors will exhibit maximal non-Gaussianity.
Maximization of non-Gaussianity has a connection to information theory as well: it is equivalent to 
maximizing the mutual information between the observed data and the learned representation \cite{Bell1995AnDeconvolution,hyvarinen1999fast}, 
and forms the basis of a group of models called Independent Components Analysis or \textsc{ICA}.
There are many variants of \textsc{ICA}, including the original neural networks approach \cite{herault1986space},
information-theoretic approaches \cite{Bell1995AnDeconvolution, hyvarinen1999fast}, and
statistical approaches \cite{amari1998adaptive}. \textsc{ICA} models, like \textsc{PCA}, generally assume a linear relationship
between observation and representation, but there has been some research into developing
non-linear \textsc{ICA}-like methods \cite{taleb1999source}.

\subsubsubsection{Sparse ICA}
\gen{Generative Assumptions.}\quad 
Many factors found in naturalistic environments are sparsely distributed. For example, consider the set
of photographs of objects. There are many possible objects, but most photographs will only display
a small number of them, and any given object will only occur in a relatively small number of photographs.
Sparsity can be interpreted in two ways --- only a limited number of factors are present for any given
observation, and for any factor, it is present in only relatively few observations.
Of course, not all components in naturalistic images are sparse: For example, the mean 
luminance of images tends to follow a roughly uniform distribution, which is sub-Gaussian. 
However, sparsity is well-matched to most features of naturalistic images relevant to tasks 
such as object classification.

\ind{Inductive Bias.}\quad 
Environments with sparse factors offer a nice opportunity to employ a stronger independence bias.
Sparse distributions are a type of non-Gaussian distribution called a \emph{super-Gaussian} distribution.
These distributions have higher kurtosis (fourth moment) than a Gaussian. Examples
of super-Gaussian distributions include the Laplacian (used in L1/lasso regularization) and Cauchy 
distributions. Sparsity is also used in the field of sparse coding to find unique solutions to over-complete
problems. We can transform \textsc{PCA} into \textsc{ICA} by adding a sparsity penalty $s(\mathbf{z})$ to the loss function
from Equation~\ref{eq:pcaloss}:
\begin{equation}\label{eq:icaloss}
\mathcal{L}_{\mathrm{ICA}} =  \lVert  \mathbf{x - \tilde{x}}  \rVert_2^2 + \lambda~ s(\mathbf{z})
\end{equation}
$\lambda$ is a parameter governing the loss trade-off between reconstruction and sparsity. 

The sparsity penalty corresponds to minimizing the negative log of the probability of \textbf{z} under some super-Gaussian
prior in a bayesian setting. For example, let's assume $s(\mathbf{z})$ corresponds to an L1 penalty, $\lVert \mathbf{z} \rVert_1$.
The L1 penalty is equivalent to a super-Gaussian Laplacian prior.  First, we assume independence of the factors, so 
the prior $p(\mathbf{z})$ can be factorized  as:
\begin{equation}
p(\mathbf{z}) = \prod_{i}{p(\mathbf{z_i})}
\end{equation}
The Laplacian prior probability density is typically defined as\footnote{For simplicity, I have omitted the scaling factor}:
\begin{equation}\label{eq:laplace}
    p_{\mathrm{Laplace}}(\mathbf{z}) \propto \prod_i{ \exp{ - \lambda \lvert \mathbf{z}_i \rvert } }
\end{equation}
This laplace prior is centered/peaked at zero (if it were not, then the resulting distribution would not be sparse), 
and $\lambda$ represents a shape parameter and is equivalent to the parameter governing the loss tradeoff.
For the purpose of optimization, we often minimize the negative logarithm of Equation~\ref{eq:laplace}, 
which leads us to the formula for $s(\mathbf{z})$:
\begin{equation}
\lambda~s(\mathbf{z}) = \sum_i{ \lambda \lvert \mathbf{z}_i \rvert } = \lambda~\lVert \mathbf{z} \rVert_1
\end{equation}

The formulation of \textsc{ICA} in Equation~\ref{eq:icaloss} is called Reconstruction \textsc{ICA} \cite{Le2011ICALearning}. 
The reconstruction loss in Equation~\ref{eq:icaloss} can also be replaced with a restriction that the representation be 
fully invertible, and helps to prevent degenerate solutions. 
Of course not all \textsc{ICA} methods use sparse factors --- some methods explicitly seek sub-Gaussian 
factors, or some other non-Gaussian characteristic. Likewise, not all methods that employ sparsity 
can be reduced to ICA; for example, sparse autoencoders are not \textsc{ICA} because they allow for
non-linear relationships between observation and representation. In the next section, I will describe
some of these non-ICA methods for non-Gaussian factors.

\rep{Representation Structure.}\quad The representation of \textsc{ICA} is exactly like that of \textsc{PCA} ---
each observation is mapped to a vector of factor coefficients.

\subsubsubsection{Non-ICA}
There are a few other methods for finding non-Gaussian independent factors that are not \textsc{ICA}, in that
they do not explicitly maximize non-Gaussianity of factors. 
Like \textsc{Sparse ICA}, these algorithms all make some assumption about the environment; for example, they
assume that the factors are binary, or make some assumption about the generative
process of the environment.

We can incorporate the bias of a sparse representation into an autoencoder architecture
by adding a sparsity penalty \cite{Makhzani2015Winner-Take-AllAutoencoders} to get 
a sparse autoencoder.

In \cite{Schmidhuber1992LearningMinimization}, binary factorial representations are learned.
In this paper, the reconstruction error is minimized alongside an 
additional penalty term: the predictability of a factor given 
all the other factors. This penalty term is made tractable by the 
assumption that all representations have binary coefficients.
This penalty works in non-Gaussian settings because it looks at dependencies
beyond correlations.

A cooperative vector quantizer 
\cite{Zemel1993ALearning,Ghahramani1995FactorialAlgorithm}
is composed of several vector quantizer units, each of which make a weighted contribution
to the output. The vector quantizer can be viewed as implementing sparsity --- only one unit
within the VQ can be active at any time. The combination of outputs from several vector 
quantizers models an environment in which several groups of sparse factors contribute
to each observation.

In \cite{Saund1995ALearning}, the author found that factorial representations can be discovered
in non-sparse settings by designing the mapping function to discourage over-cooperation between
factors, and to closely reflect the generative structure of the environment.

Austerweil and Griffiths \cite{Austerweil2009TheLearning} investigate how humans might
select features to use for category learning. They use an Indian Buffet Process prior, which
corresponds to an assumption that humans will learn independent features --- features that covary 
will be grouped together. They find that their model predicts human categorization behavior
in experiments.

In the next section, I will compare the bias of Gaussian-distributed factors with the bias of
sparsely distributed factors on a few example sparse datasets.

\subsubsection{Comparison of Gaussian and Sparsity Bias in a Sparse Environment}
The relationship of \textsc{PCA} and \textsc{ICA} can be seen in the model organization chart in Figure~\ref{fig:model_chart}.
Both \textsc{PCA} and \textsc{ICA} have fairly high inductive bias due to their linear mapping assumption.
The difference in inductive bias between the two models is obvious when
comparing equations~\ref{eq:pcaloss} and ~\ref{eq:icaloss} --- \textsc{ICA} has an extra
term (sparsity) in the objective.

\begin{figure}[ht]
\center
\begin{subfigure}[b]{0.3\textwidth}
\includegraphics[scale=0.3]{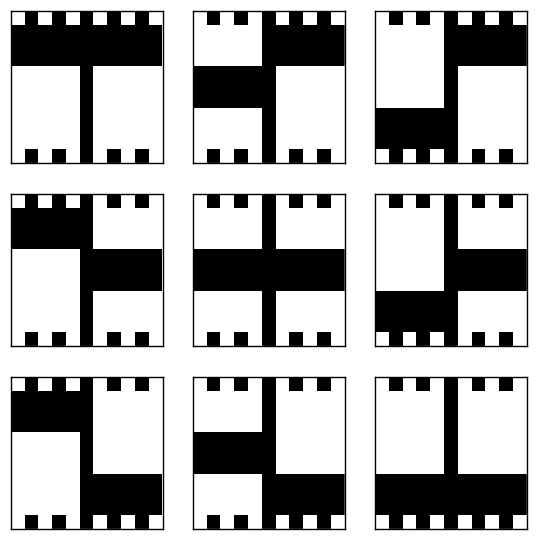}
\end{subfigure}
~
\begin{subfigure}[b]{0.6\textwidth}
\centering
\includegraphics[scale=0.15]{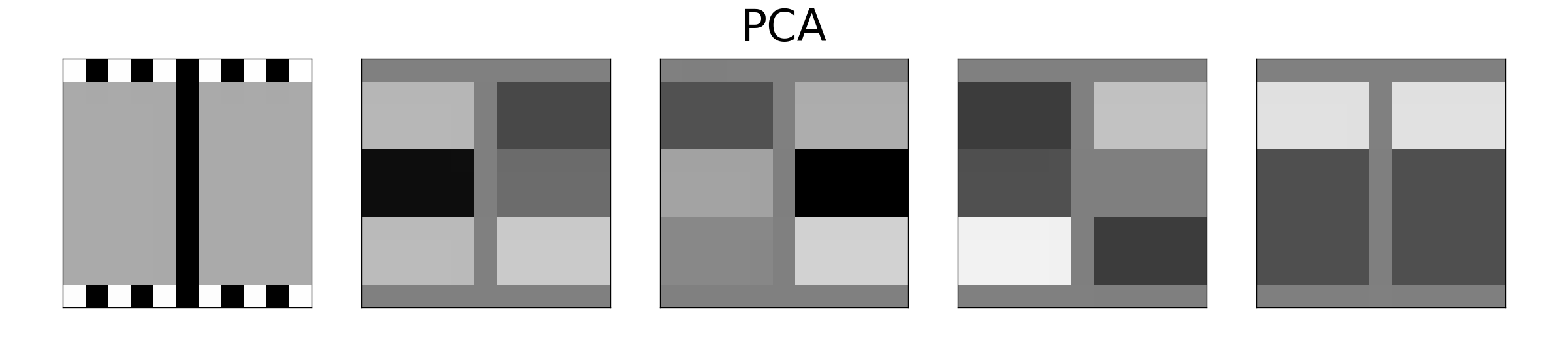}
\includegraphics[scale=0.15]{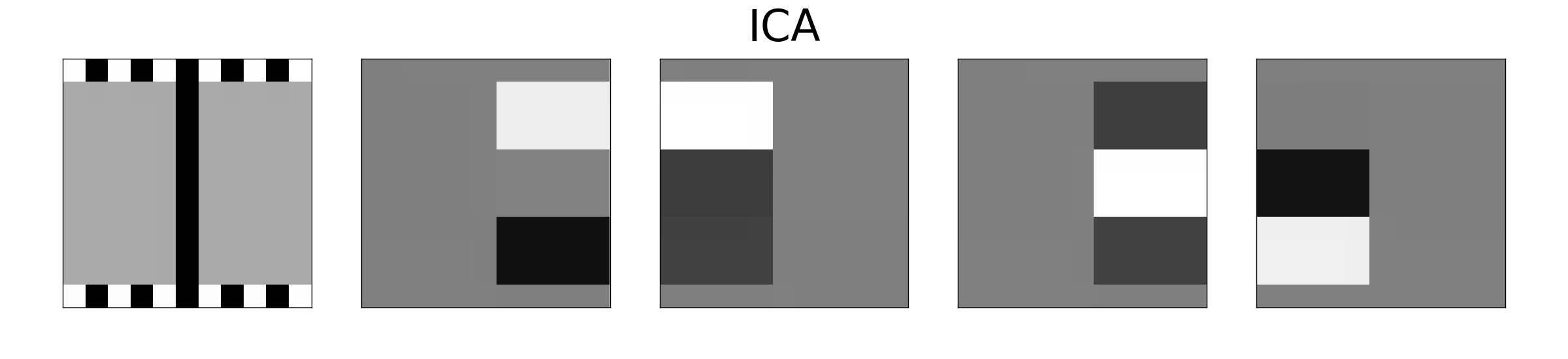}
\end{subfigure}
\caption{(left) A synthetic dataset like the one from \cite{Saund1995ALearning}. Nine 121-dimensional test data samples, where
the positions of the black rectangles on the left and right sides is controlled by two independent processes.
(top right) The mean vector and four components learned by a \textsc{PCA} model.
(bottom right) The mean vector and four components learned by an \textsc{ICA} model.
}
\label{fig:saund1995_1}
\end{figure}

A problem with \textsc{PCA}/decorrelation bias is that is not well-matched to environments where the independent
components are sparse.
An illustration of the problem can be found in 
\cite{Saund1995ALearning}, which I have recreated here in Figure~\ref{fig:saund1995_1} on the left. The synthetic dataset
consists of ($11\times 11$)-dimensional vectors of numbers, where each number is either on (1, white) or off (0, black). 
One factor controls the position of the black rectangle on the left, and a separate, independent, factor controls
the position of the black rectangle on the right. The factors in this synthetic environment are sparse --- only 
one rectangle on each side is active at a time. Also, the dataset is fully factorial, showing all combinations
of positions once each.
I trained both a \textsc{PCA} and \textsc{ICA} model on this dataset.
At the top right in Figure~\ref{fig:saund1995_1}, we see the
mean vector (left) and components learned by \textsc{PCA} --- the rows of the $\mathbf{W^{-1}}$ matrix. In this visualization,
the pixels are scaled so that grey represents a neutral (zero) weight, black represents a negative coefficient, and 
white represents  a positive coefficient.
The basis learned by \textsc{PCA} is compact and faithful --- 
only four factors (plus the mean vector) can explain all the variation in the dataset. 
However, the representation is not interpretable --- the generative process that synthesized the dataset
is not at all apparent in the representation. 
In other words, the bias of decorrelation is insufficient
in explaining the true generative process of the environment. At the bottom right in Figure~\ref{fig:saund1995_1},
we see an \textsc{ICA} decomposition of the same dataset\footnote{Using the FastICA\cite{hyvarinen1999fast} algorithm}. 
The components shown are again the rows of the $\mathbf{W^{-1}}$ matrix, which in the parlance of \textsc{ICA} is called
the "mixing matrix".
The four \textsc{ICA} components reflect the causal structure of this environment better --- each 
component only describes activity on one side of the image. The \textsc{ICA} representation is just as compact
and faithful as the \textsc{PCA} representation since all original examples can be reconstructed from the four
components.

\begin{figure}[ht]
\center
 \includegraphics[scale=0.25]{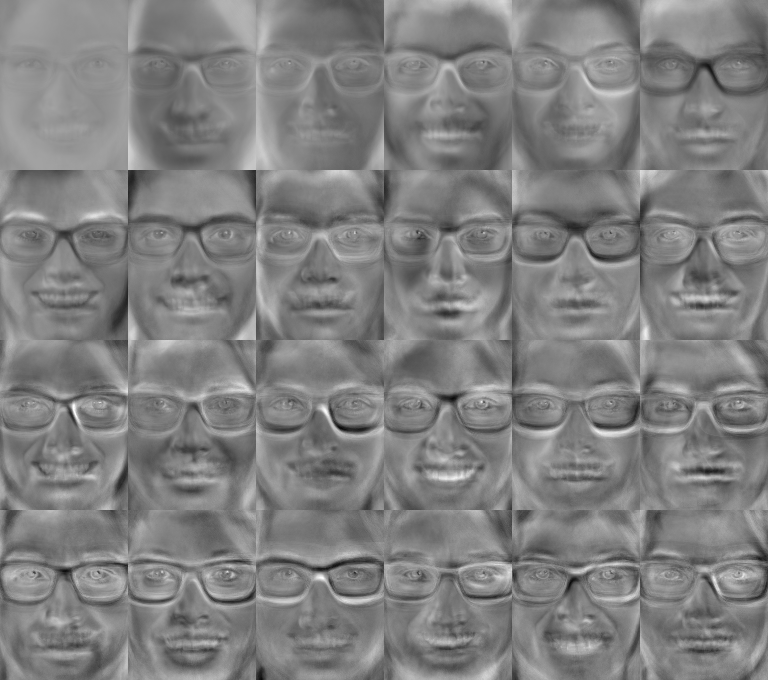}
 \includegraphics[scale=0.25]{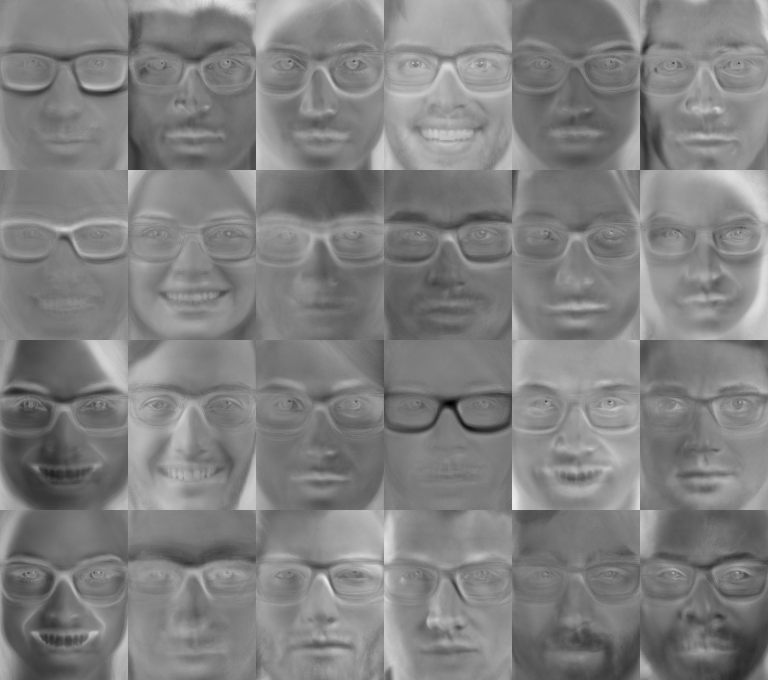}
\caption{(left) The top 24 components for the glasses data set. 
(right) 24 \textsc{ICA} components for the glasses dataset.}
\label{fig:framesdirect_pca_ica}
\end{figure}

Another example where \textsc{ICA} improves interpretability over \textsc{PCA} is shown in Figure~\ref{fig:framesdirect_pca_ica},
using the naturalistic face dataset from the introduction. Once again, I trained both a \textsc{PCA} 
and \textsc{ICA} model on the dataset. On the left are the 
24 top components (again, rows of $\mathbf{W^{-1}}$, scaled in the same way as Figure~\ref{fig:saund1995_1}) 
learned by \textsc{PCA}, and on the right 24 components learned by \textsc{ICA}.
In this dataset, there are several independent
sources of variation including identity, lighting conditions, and glasses style. Subjectively, we can see that 
each \textsc{PCA} component seems to highlight a mixture of person identity, glasses style, and lighting conditions.
The \textsc{ICA} components tend to focus on one cause --- if glasses are emphasized then identity and lighting
conditions are de-emphasized. It is also notable that unlike \textsc{PCA}, \textsc{ICA} does not produce an 
ordered list of components in terms of how much variance they explain. Note that the \textsc{ICA} components do
not isolate the causal factors perfectly --- \textsc{ICA} is still limited by the inductive bias of the linear
mapping between observation and representation, so it cannot disentangle glasses from the rest of
the face.

\begin{figure}[ht]
\center
\raisebox{0.25\height}{\includegraphics[scale=0.3]{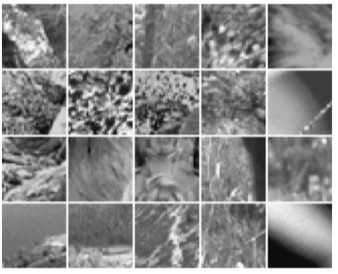}}
\includegraphics[scale=0.2]{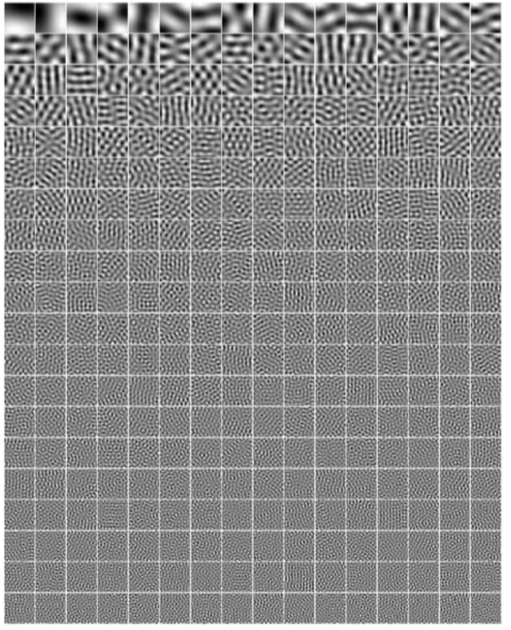}
\raisebox{0.1\height}{\includegraphics[scale=0.2]{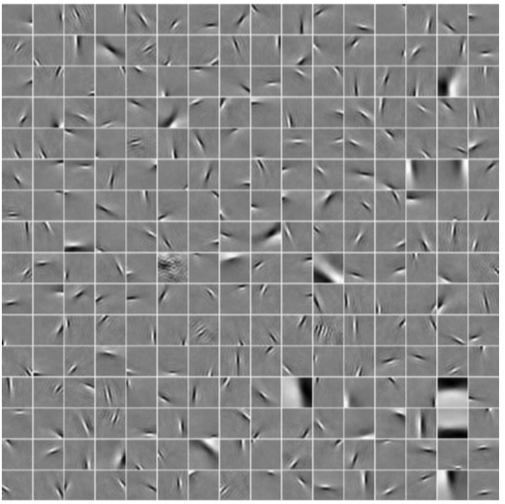}}
\caption{(left) Samples from a data set of nature images. Results are from \cite{hyvarinen2009natural}.
(center) A visualization of the \textsc{PCA} basis of this dataset.
(right) A visualization of an \textsc{ICA} basis for this dataset. }
\label{fig:hyvarinen_pca_ica}
\end{figure}

A less constrained environment is shown in Figure~\ref{fig:hyvarinen_pca_ica}, which shows samples from a 
dataset of nature images from \cite{hyvarinen2009natural}. Each image is
split into $32\times32$ patches, and the patches were used to train \textsc{PCA} and \textsc{ICA} models. In this environment,
each image patch can be interpreted as having been generated by a number of localized factors that operate
independently at different scales --- the crest of a wave is generated independently from the overall lighting
of the scene. In the center is the \textsc{PCA} basis and on the right is the \textsc{ICA} basis, with components scaled the same way as
the other figures. These images are all from \cite{hyvarinen2009natural}.
The \textsc{PCA} components all code for global features at different scales. The \textsc{ICA} components also seem
to operate at different scales, but are much more localized, better matching the generative process
of the environment.

I have shown how the bias of factorial representations can help learn better representations
given a factorial environment.
For sparse environments observed in naturalistic images, decorrelation is an insufficient
bias to learn factorial representations. The stronger bias of \textsc{ICA} with sparsity is shown to improve representations.
\textsc{ICA} solutions are also unique, which make them more interpretable than non-unique \textsc{PCA} solutions.
Next I will describe a bias that assumes factor representations
should be insensitive to variations in their expression that are unrelated
to the factor.
\subsection{Invariance Bias}
Factors can vary in how they are expressed. Objects can appear at 
various locations in an image, or be rotated and scaled. Linear methods
like \textsc{PCA} and \textsc{ICA} are unable to express this type of variation, and tend to learn
redundant representations of factors to describe all the ways in which a factor can be realized.
In these environments, we would prefer a representation of a factor to be \emph{invariant} to
random fluctuations in how that factor is expressed.

\gen{Generative Assumptions.}\quad Many factors can be expressed in the input in
more than one way.
In a nature scene, animals might appear in different parts of the image.
It would be nice to be able to represent the presence of the animal as one coherent factor, rather than
have a code for each animal in each location. 
Likewise, a texture pattern in an image might show up in various phases, or under rotations.
In the introductory example of shapes and colors, the "shape" factor is invariant to color, and vice versa.

I use the term \emph{view} to refer to one of the possible realizations of a factor ---
e.g. one location, rotation angle, or one translation.
Views of a factor can \emph{compete} or \emph{cooperate} to generate an observation.
For example, there is competition between locations for an animal to appear in the nature
scene: the locations are mutually exclusive. 
However, views could exist at only the extremes of locations, and observations would
interpolate between these views. In this case, the views cooperate, and are not mutually exclusive.

I will first show an inductive bias that that is consistent with an environment where
factors compete, and views within a factor cooperate. 
Then I will show an inductive bias consistent with an environment
where factors cooperate and views within a factor compete.

\subsubsection{ISA}
\ind{Inductive Bias.}\quad A method for learning invariant factors, where
views cooperate is to \emph{pool} similar views together.
In the case of \textsc{ISA}, a view corresponds to a linear component, of the same
dimensionality as the input, just like in \textsc{PCA} and \textsc{ICA}.
The magnitude of all the coefficients in the pool is the coefficient on the factor. 
Sparsity is then applied at the pooled layer, and has the effect of making the pools
compete with each other.
The subspace formed by the pool represents all the views of the factor.
Independent Subspace  Analysis (\textsc{ISA})\cite{hyvarinen2000emergence} is an extension of 
\textsc{ICA} that implements this pooling method. 
In the generative model of \textsc{ISA}, we first choose some small number of factors to be present and then 
generate the observation by linearly combining elements from within the pools representing the
active factors.

\begin{figure}[ht]
\center
\begin{tikzpicture}[scale=1.3]
    \draw (1,1) circle (0.2cm);
    \draw (1,2) circle (0.2cm);
    \draw (2,1) circle (0.2cm);
    \draw (2,2) circle (0.2cm);
    \draw[rounded corners=10pt] (0.5,0.6) rectangle ++(2,0.75);
    \draw[rounded corners=10pt] (0.5,1.6) rectangle ++(2,0.75);
    \node [right] at (2.75,2.) {$\mathbf{\hat{z}}_1 = \sqrt{\mathbf{z}_1^2 + \mathbf{z}_2^2}$};
    \node [right] at (2.75,1.) {$\mathbf{\hat{z}}_2 = \sqrt{\mathbf{z}_3^2 + \mathbf{z}_4^2}$};
    \node [] at (1,1) {\small $\mathbf{z}_3$};
    \node [] at (1,2) {\small $\mathbf{z}_1$};
    \node [] at (2,1) {\small $\mathbf{z}_4$};
    \node [] at (2,2) {\small $\mathbf{z}_2$};
\end{tikzpicture}
\caption{ Illustration of the grouping structure of \textsc{ISA}. The circles represent the coefficients z. }
\label{fig:isa_illustration}
\end{figure}
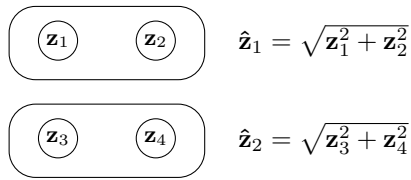

In this model, there are two layers --- the first is a linear layer, and the second
pools the linear components together to form the final representation.
The first layer of $\mathrm{N}$ coefficients $\mathbf{z}$ is, like \textsc{ICA}, linearly related to the input $\mathbf{x}$:
\begin{equation}
    \mathbf{z} = \mathbf{xW}
\end{equation}
A grouping function $g(\mathbf{z})$ pools together some subset of $\mathbf{z}$, in this case by their
Euclidean norm:
\begin{equation}\label{eq:isa_sum}
    g(\mathbf{z_{...}}) = \sqrt{ \sum_i{ \mathbf{z}_i^2} }
\end{equation}
The second layer, the set of pooled components, is called $\hat{\mathbf{z}}$. In this case,
$\hat{\mathbf{z}}$ is composed of $N/D$ pools of size $D$, but the size and number of pools is arbitrary:
\begin{equation}
    \hat{\mathbf{z}} = [ g(\mathbf{z}_{1..D}) , g(\mathbf{z}_{D+1..2D}), ~...~, g(\mathbf{z}_{(N-1)D..ND}) ]
\end{equation}
The loss function of \textsc{ISA} is identical to that of \textsc{ICA}, except that the sparsity loss $s(\cdot)$
is applied to $\hat{\mathbf{z}}$ instead of $\mathbf{z}$:
\begin{equation}
\mathcal{L}_{\mathrm{ISA}} =  \lVert  \mathbf{x - \tilde{x}}  \rVert_2^2 + \lambda~s(\hat{\mathbf{z}})
\end{equation}
Again, $\lambda$ governs the trade-off between reconstruction loss and the sparsity penalty.
The grouping structure of \textsc{ISA} is illustrated in Figure~\ref{fig:isa_illustration}.

Sparsity is applied at the pooled layer $\mathbf{\hat{z}}$, so only the pools, not the 
individual components within a pool,
are encouraged to be sparse. This has the side-effect of encouraging the components of each pool
to be \emph{more dependent} on each other, and increase cooperation. 
Intuitively, if the components of a pool are independent,
then their sum in Equation~\ref{eq:isa_sum} should be more Gaussian, according to the central
limit theorem. The sparsity objective eliminates this possibility, and so encourages the
components in the pool to be more dependent. 

\rep{Representation Structure.}\quad 
\textsc{ISA} has a more complex structure than \textsc{ICA}, in that it has two layers. Each factor in
the second layer $\mathbf{\hat{z}}$ is accompanied by a vector of coefficients in its corresponding
\emph{subspace} in $\mathbf{z}$. The representation is therefore composed of both a scalar coefficient in $\mathbf{\hat{z}}$ as well
as the coordinate in its corresponding z subspace.

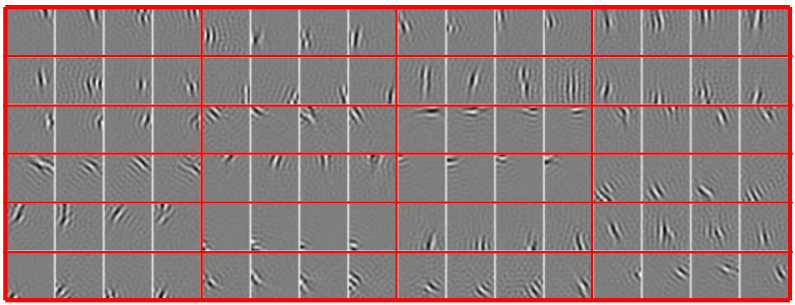
\begin{figure}[ht]
\centering
\begin{tikzpicture}
    \node[anchor=south west,inner sep=0] at (0,0) {\includegraphics[scale=0.5,trim={0 13.1cm 0 0},clip]{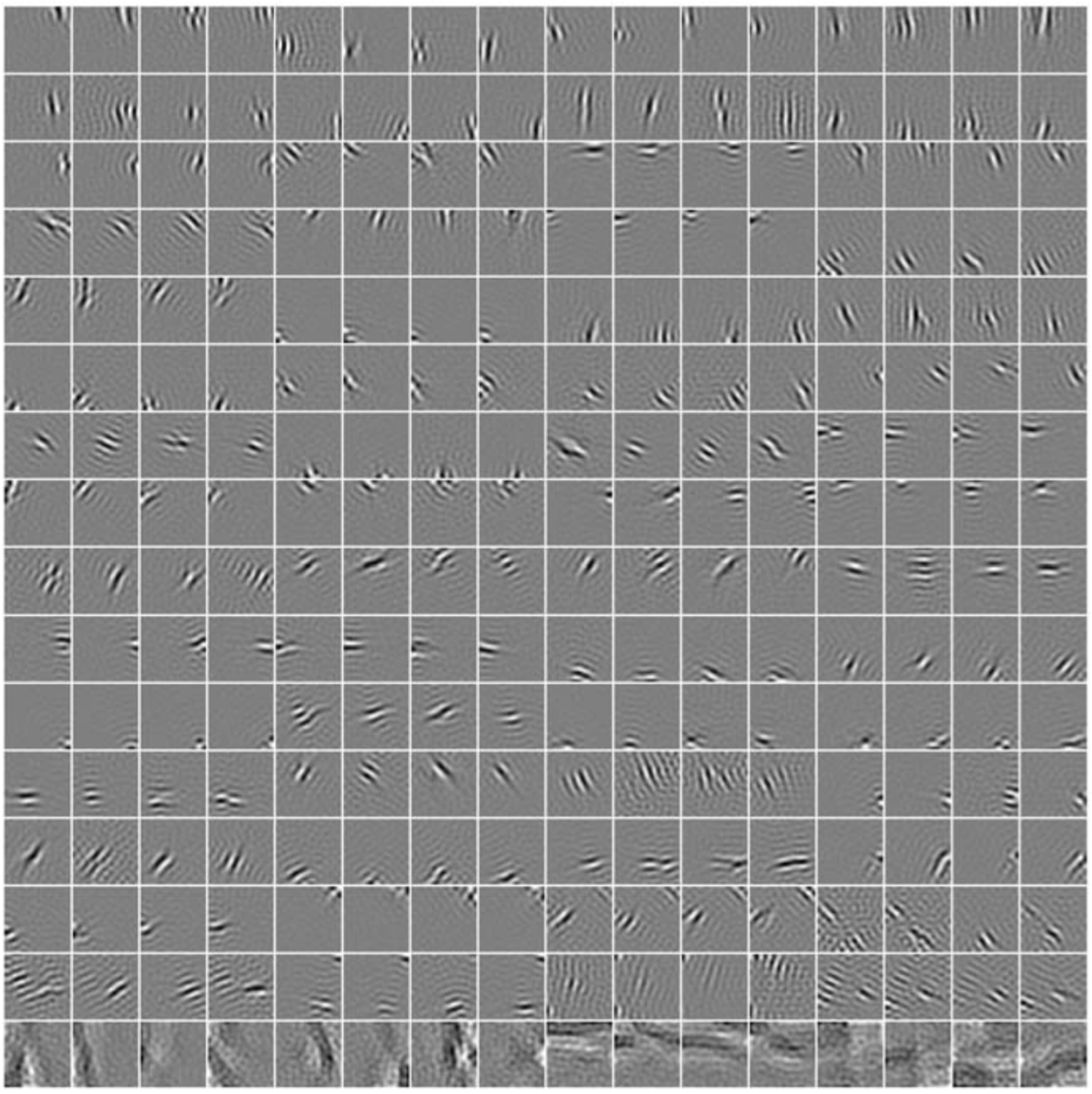}};
    \draw[red,thick] (2.64,0) -- (2.64,3.9);
    \draw[red,thick] (5.23,0) -- (5.23,3.9);
    \draw[red,thick] (7.83,0) -- (7.83,3.9);
    
    \draw[red,ultra thick] (0.03,0) -- (0.03,3.9);
    \draw[red,ultra thick] (10.46,0) -- (10.46,3.9);
    \draw[red,ultra thick] (.03,0) -- (10.46,0);
    \draw[red,ultra thick] (.03,3.9) -- (10.46,3.9);
    
    \draw[red,thick] (0,0.64) -- (10.5,0.64);
    \draw[red,thick] (0,1.30) -- (10.5,1.30);
    \draw[red,thick] (0,1.95) -- (10.5,1.95);
    \draw[red,thick] (0,2.59) -- (10.5,2.59);
    \draw[red,thick] (0,3.25) -- (10.5,3.25);
\end{tikzpicture}
\caption{Components learned by independent subspace analysis for the dataset of
32$\times$32 patches of naturalistic images from \cite{hyvarinen2009natural}. 
Every four consecutive components in a row corresponds to one pooled component. Positive weights are
shown as light regions, and negative weights are shown as dark, while weights close to zero are gray.}
\label{fig:isa_hyvarinen}
\end{figure}

\examp{Examples.}\quad 
A visualization of some \textsc{ISA} pools, from \cite{hyvarinen2009natural}, is shown in 
Figure~\ref{fig:isa_hyvarinen}. The authors applied \textsc{ISA} to the same dataset of nature 
images that they used for \textsc{ICA} in Figure~\ref{fig:hyvarinen_pca_ica}. 
A visualization of the learned components inside the $\mathbf{W^{-1}}$ matrix is 
shown in Figure~\ref{fig:isa_hyvarinen}.  In this case, the number of components 
per pool was set to four, and the pools are ordered so that the sparsest ones, according
to $s(\cdot)$, are at
the top. The analysis in \cite{hyvarinen2009natural} indicates that the pools are less
sensitive to phase than features learned by \textsc{ICA}, meaning that in this case,
the factors in \textsc{ISA} have more phase-invariance.
In an \textsc{ICA} solution, the components are highly localized in space and operate
at different scales (e.g. Figure~\ref{fig:hyvarinen_pca_ica}).
In \textsc{ISA}, each pool's components are at the same scale, localized within
the same region of the image, and seem to be oriented in the same way. 

There are, of course, many other types of inductive bias for invariant features,
especially for image processing, where there are known classes of variation. 
Models often have hard-wired invariance for these types of variations.
These biases are helpful for learning in their respective domains, and 
are used in many of the models in later sections. However, they are
domain-specific adaptations: they are more concerned with specifics of the \emph{environment} than
with the \emph{representation}. Therefore, their purpose is largely orthogonal to the purpose of 
this work, which is to describe how to learn factorial representations in any environment.

The inductive bias of \textsc{ISA} is weaker than \textsc{ICA} because it introduces a non-linearity
into the mapping function, and reduces the number of components that sparsity is applied to.
\textsc{ISA} can be reduced to \textsc{ICA} by setting the number of groups equal to the number of
first-layer components.
Therefore, it appears to the left of \textsc{ICA} on the model chart in Figure~\ref{fig:model_chart}.

Next, I will demonstrate an inductive bias that is appropriate for an environment in which factors 
cooperate with each other and components within a factor compete.

\noindent \textbf{Other Methods With Cooperative Pools.}\quad
Factor models with cooperative pools are used in a variety of contexts.
Pooling can be used in an overlapping spatial grid to create a topographic map 
representation, as in Kohonen's adaptive-self organizing maps 
\cite{kohonen1996emergence}. Topographic maps lay out the first level of the
latent representation as a grid. Pooling happens in overlapping sub-regions
of this spatial grid. This encourages each sub-region to be more similar,
so the filters in each grid cell have spatial coherency.
A topographic map model incorporating sparsity can be found in \cite{Kavukcuoglu2009LearningMaps}.

\subsubsection{Cooperative Vector Quantizer}
\ind{Inductive Bias.} In some environments, views compete and factors cooperate.
In \cite{Zemel1993ALearning}, each factor is represented by a Vector Quantizer (VQ),
which corresponds to a pool of views that compete to explain the observation.
An array of VQs cooperate to generate the output. Each VQ is composed of a pool of linear
components, just like in \textsc{ISA}.
This architecture is called a Cooperative Vector Quantizer (\textsc{CVQ}).
Unlike ISA, in the \textsc{CVQ}, the components in the pool are mutually exclusive:
only one of them may be active at a time. Additionally, there is no competition imposed
between the factors. They are all assumed to be present all the time.

The inductive bias of \textsc{CVQ} is the same as \textsc{ISA}. The difference is only
in whether sparsity/competition is applied within each pool or on the pools themselves.

\rep{Representation Structure.} The representation structure of \textsc{CVQ} is identical
to that of \textsc{ISA}: each factor is represented by a pool of components, each of
which is linearly related to the input.

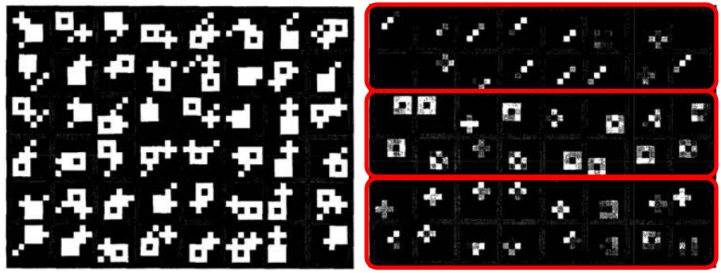
\begin{figure}[ht]
    \centering
    \includegraphics[scale=0.3]{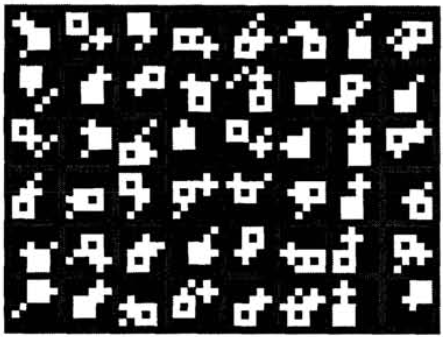}
    \begin{tikzpicture}
        \node[anchor=south west,inner sep=0] at (0,0) {\includegraphics[scale=0.3]{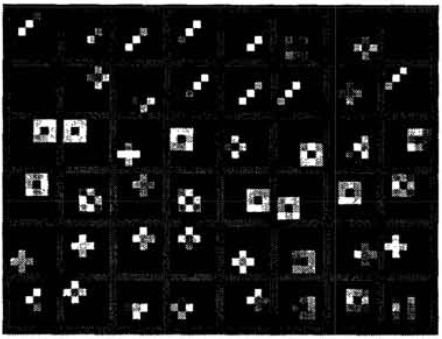}};
        \draw[red,ultra thick,rounded corners] (0,0.05) rectangle (4.65,1.24);
        \draw[red,ultra thick,rounded corners] (0,1.26) rectangle (4.65,2.4);
        \draw[red,ultra thick,rounded corners] (0,2.4) rectangle (4.65,3.55);
    \end{tikzpicture}
    \caption{(left) Samples from the shapes dataset used for training in \cite{Ghahramani1995FactorialAlgorithm}.
    (right) A representation learned by \textsc{CVQ}, also from \cite{Ghahramani1995FactorialAlgorithm}.
    Each red box corresponds to one pool of a VQ. }
    \label{fig:cvq_example_1}
\end{figure}

\examp{Examples.} In \cite{Ghahramani1995FactorialAlgorithm}, a \textsc{CVQ} is trained
on a dataset of overlapping shapes, reproduced here in Figure~\ref{fig:cvq_example_1}.
In the dataset, each $6 \times 6$ black-and-white image is generated by combining three 
white shapes: a cross, a diagonal line, and an empty square, in a number of different locations. 
Most examples have all three shapes, although in some the shapes are completely overlapping.
Therefore, this example has three cooperating factors, and each factor is realized in only
one component.
The \textsc{CVQ} was trained with three pools/VQs.
The components of both pools are shown on the right in Figure~\ref{fig:cvq_example_1}. 
The first pool clearly represents the diagonal line, in various locations in the image.
The second represents the empty square, and the last one represents the cross. 
In this example, the model has learned to represent shape in a way that is invariant to
location.

\subsubsection{Other Inductive Biases for Invariance}
In \cite{austerweil2010learning}, a transformed Indian Buffet Process prior is applied to 

Some models encourage representations to be invariant to minor perturbations in the input. A contractive
regularization penalty can be used to encourage factors to be insensitive to 
local directions of variation in the observation space \cite{Rifai2012DisentanglingRecognition}. 
Denoising autoencoders \cite{Vincent2008ExtractingAutoencoders} also encourage
representations to invariant to noise added to the observation. 

Next, I will describe a setting in which factors can vary in how they are expressed, but it
is the interaction between factors that is responsible for generating observations.

\subsection{Combination Bias}
I will highlight three different modeling biases that determine how
factors can interact in generating observations. 
The simplest and strongest inductive bias for factor combination is the \emph{linear} bias.
In a linear model, observations are generated through a weighted summation of factor components.
The coefficients can be either negative or positive, so a factor can cancel
the contributions of other factors. \textsc{PCA} and \textsc{ICA} are both examples of
this linear bias.
First, I will discuss a bias that allows for \emph{multiplicative interactions} to occur
between a small set of factors. 
Environments where factors are mutually exclusive in explaining portions
of the observation are called \emph{functional parts}; I will describe
some biases that work well in these environments.
Finally, some environments are composed of a \emph{hierarchy of layers}
of factors, which generate observations at increasing levels of detail
and decreasing levels of abstraction.

\subsubsection{Multilinear}
\gen{Generative Assumptions.}\quad 
In some environments, a small number of factors are known to \emph{always} be present, and
their interactions are responsible for generating observations. 
Each factor has a number of views, and
view interacts with the views of the other factors.
For example, the face dataset from the introductory section can be expressed
as the interaction of person identity and glasses style --- the interactions between these two 
factors can account of a large portion of the variation in images. A person's identity modulates
where the glasses sit on the face, and the glasses style modulates the shadow cast on the
person's face.

\ind{Inductive Bias.}\quad 
These environments are the motivation behind the general class of \textsc{Multilinear} models,
in which the multiplicative interactions between factors account for the variability of
the observations. First, I will describe the simplest case of a \textsc{Multilinear} model, called
a \emph{bilinear} model, in which there are only two factors. I will describe how \textsc{Multilinear} 
models can be combined with an \textsc{ICA}-like sparsity penalty. Finally, I will show an example
with three interacting factors. 

\subsubsubsection{Bilinear Models}
In a bilinear model \cite{Tenenbaum2000SeparatingModels.}, two factors interact with each other to generate
each observation. Like \textsc{ICA} and \textsc{CVQ}, each factor in a bilinear model
is represented by a pool of coefficients. 
Each coefficient of a factor interacts multiplicatively with the coefficients from the other factor.
Each combination of coefficients is associated with a component vector, of the same
dimensionality as the input.
The weight on a component is determined by the product of the two coefficients associated with it,
and the observation is generated by a summation of these weighted components.
In \cite{Tenenbaum2000SeparatingModels.}, these factors are named "content" and "style", but
they can be any two factors. 
For example, consider the environment of upright characters and italicized characters.
The appearance of a character is completely determined by the interaction of two factors: the identity
of the character, and its style. 

Each factor in a bilinear model is represented by a "pooled" structure like in \textsc{ISA}.
Suppose we are modeling two factors, a and b, with representations $\mathbf{z^a}, \mathbf{z^b}$.
There are $M$ components in the pool $\mathbf{z^a}$, and $N$ components in pool $\mathbf{z^b}$.
The weight matrix $\mathbf{W}$ is three-dimensional ($N \times M \times D$), and each 
component of W is weighted by the interaction between $\mathbf{z^a}$ and $\mathbf{z^b}$:
\begin{equation}
    \mathbf{\tilde{x}} = \sum_i^M{ \sum_j^N{ \mathbf{W}_{ij} \mathbf{z}^\mathbf{a}_i \mathbf{z}^\mathbf{b}_j } }
\end{equation}
As in \textsc{PCA}, bilinear models minimize reconstruction loss.
We can interpret this model as a \textsc{PCA} model where the representation $\mathbf{Z}$
is a matrix that is factored into the interactions between the two vectors $\mathbf{z^a}$ and $\mathbf{z^b}$: 
$\mathbf{Z = z^a} (\mathbf{z}^\mathbf{b})^\mathrm{T}$.
This model is essentially conjunctive: for a given
coefficient $\mathbf{Z}_{ij}$ to be non-zero, both $\mathbf{z}^\mathbf{a}_i$ and $\mathbf{z}^\mathbf{b}_j$
have to be non-zero. 

\begin{figure}[ht]
\center
\begin{tikzpicture}[scale=1.3]
    \draw (1,1) circle (0.2cm);
    \draw (1,2) circle (0.2cm);
    \draw (2,1) circle (0.2cm);
    \draw (2,2) circle (0.2cm);
    \draw[rounded corners=10pt] (0.5,0.65) rectangle ++(2,0.7);
    \draw[rounded corners=10pt] (0.5,1.65) rectangle ++(2,0.7);
    \draw[rounded corners=10pt] (0.65,0.5) rectangle ++(0.7,2);
    \draw[rounded corners=10pt] (1.65,0.5) rectangle ++(0.7,2);
    
    \node [left] at (1.25,2.75) {$\mathbf{z}^\mathbf{a}_1$};
    \node [left] at (2.25,2.75) {$\mathbf{z}^\mathbf{a}_2$};
    \node [right] at (2.7,2.) {$\mathbf{z}^\mathbf{b}_1$};
    \node [right] at (2.7,1.) {$\mathbf{z}^\mathbf{b}_2$};
    
    \node [] at (1,1) {\tiny $\mathbf{W}_{21}$};
    \node [] at (1,2) {\tiny $\mathbf{W}_{11}$};
    \node [] at (2,1) {\tiny $\mathbf{W}_{22}$};
    \node [] at (2,2) {\tiny $\mathbf{W}_{12}$};
\end{tikzpicture}
\caption{
Illustration of the grouping structure of a bilinear representation with two factors a and b, each
composed of two sub-factors. The circles represent the components \textbf{W}.
The weight on a component is determined by the product of the $\mathbf{z^a}$ and $\mathbf{z^b}$ coefficients of the groups 
it is a part of. }
\label{fig:bilinear_illustration}
\end{figure}
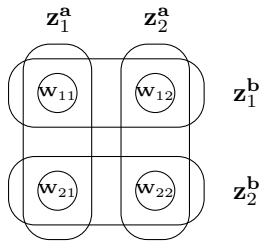

The bilinear model encourages similarity in the rows and columns of the component matrix $\mathbf{W}$. 
To help understand this effect, consider the diagram in Figure~\ref{fig:bilinear_illustration}.
The coefficient $\mathbf{z}^\mathbf{a}_1$, acts as a "gate" on the components
$\mathbf{W}_{11}$ and $\mathbf{W}_{21}$. If the $\mathbf{z}^\mathbf{a}_1$ coefficient is zero, then the entire
column is turned off. The gating property of $\mathbf{z}^\mathbf{a}_1$ encourages
$\mathbf{W}_{11}$ and $\mathbf{W}_{21}$ to both be active for the same inputs,
and thus more correlated and more similar to each other.
In this way, the bilinear model encourages similarity in both the rows and columns
of Figure~\ref{fig:bilinear_illustration}.

This model is has some characteristics that are similar to \textsc{CVQ}: each factor is
represented by a pool of linear components, and each factor is assumed to be present.

\rep{Representation Structure.}\quad 
\textsc{Multilinear} representations use pools to represent factors, like in \textsc{ICA} and \textsc{CVQ}.
However, unlike \textsc{CVQ} and \textsc{ISA}, the components associated with a pool
are encouraged be consistent from pool to pool.

\examp{Examples of Bilinear Models.}\quad 
A simple example of a bilinear model can be found in \cite{Grimes2005BilinearVision}, 
which also includes a sparsity constraint on \textbf{a} and \textbf{b}. 
Sparsity here means that, for each factor pool, we sample only a small number of components
to be present for any observation. The sparse bilinear loss function looks like:
\begin{equation}
\mathcal{L}_{\mathrm{SPARSE-BILINEAR}} = \lVert \mathbf{x - \tilde{x}} \rVert^2 + 
\lambda_1 s(\mathbf{z^a}) +
\lambda_2 s(\mathbf{z^b})
\end{equation}
The $\lambda_1$ and $\lambda_2$ settings govern the tradeoff between reconstruction
loss and the sparsity penalties on a and b.

In \cite{Grimes2005BilinearVision}, the authors compared \textsc{ICA} with a sparse bilinear model.
The dataset used for training is generated by randomly selecting 10 $\times$ 10 image
patch locations, drawn from 512 $\times$ 512 naturalistic images. 
For each image patch location, a set of image patches is generated by
shifting the patch location $\pm2$ pixels horizontally.
The authors construct a sparse bilinear model, where the image patch feature is represented
by a  pool which they name \textbf{x}, and the translation is modeled by another pool named \textbf{y}.
The model was trained using a custom training procedure that encouraged \textbf{x} to represent
image patch features, and \textbf{y} to represent translations.
This training procedure also used some \emph{supervised} bias, utilizing knowledge of the
relationship between an image feature and its translations. Supervised bias will be covered thoroughly
in a later section.
The authors also trained an \textsc{ICA} model on this dataset.

\begin{figure}[th]
\center
\includegraphics[scale=0.3]{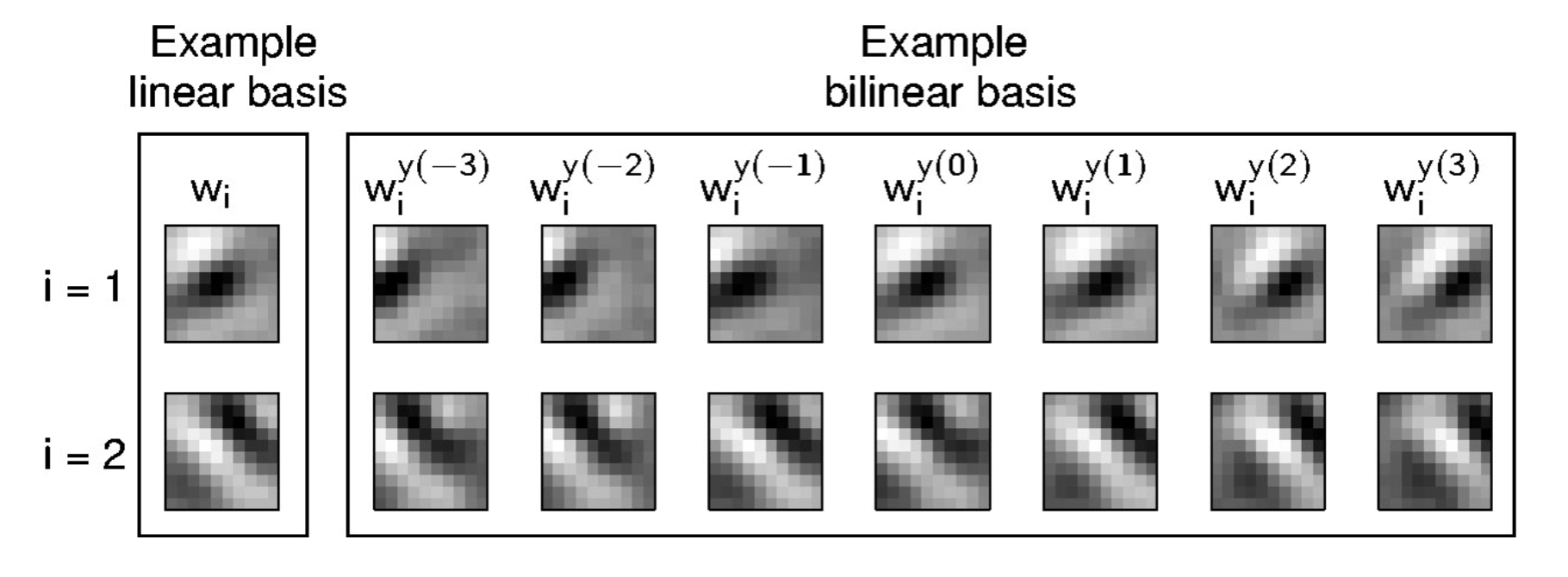}
\caption{Visualization of an example linear model, on the left, and a corresponding bilinear model, 
on the right, from \cite{Grimes2005BilinearVision}. Negative component pixels are represented as dark
regions, and positive pixels as light regions.}
\label{fig:bilinear_example_1}
\end{figure}

On the right of Figure ~\ref{fig:bilinear_example_1}, in the box labeled "Example bilinear basis",
we can see some sample components
(from the \textbf{W} matrix) corresponding to these two factors. The rows correspond to the \textbf{x} factor
(image patch feature), and the columns correspond to the \textbf{y} factor (horizontal translations).
The components in each row are consistent: they represent the same image patch feature, at different
translations. The components in each column are consistent: they represent the image patch feature
at a given translation.

On the left of Figure~\ref{fig:bilinear_example_1}, in the box labeled "Example linear basis",
two components of the \textsc{ICA} model are shown 
that are most similar to the \textbf{x} components of the examples chosen for the bilinear model.
\textsc{ICA} learns to represent similar components, but would redundantly represent different
translations of an image patch feature rather than grouping them together.

\begin{figure}[th]
\centering
\includegraphics[scale=0.3]{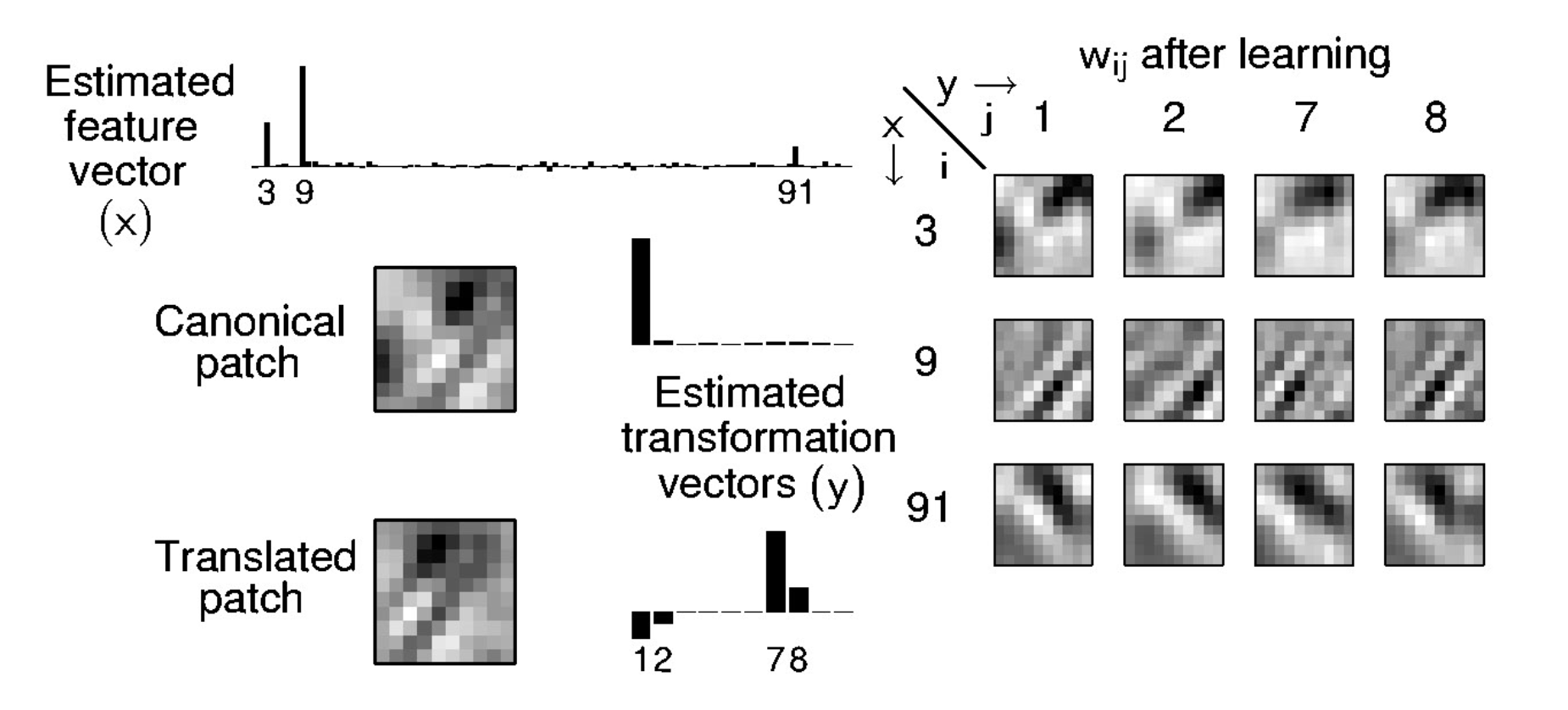}
\caption{A sample component learned by a bilinear model from \cite{Grimes2005BilinearVision}. See
text for a detailed explanation.}
\label{fig:bilinear_example_3}
\end{figure}

Figure~\ref{fig:bilinear_example_3} shows the response of \textbf{x} and \textbf{y} for a sample image patch,
labeled "Canonical patch". Note that both \textbf{x} and \textbf{y} are sparse --- very few coefficients
are significantly nonzero. According to the model, the canonical patch is a mixture of
$\mathbf{x}_3$ and $\mathbf{x}_9$ and $\mathbf{y}_1$.
A translated version of the canonical patch is also shown, labeled "Translated patch".
The \textbf{y} vector of this patch is different from the canonical patch --- instead of activating
$\mathbf{y}_1$ it activates $\mathbf{y}_{1,2,7,8}$, representing a translation.

\noindent \textbf{Other Bilinear Models.}\quad 
Not all bilinear models assume a sparse environment, or learn to represent translations. 
For example,
Tensor Analyzers \cite{Tang2013TensorAnalyzers} are bilinear models that assume 
Gaussian-distributed factors, and were applied to model lighting conditions and identity
on a face dataset.

\begin{figure}[h]
\centering
\includegraphics[scale=0.5]{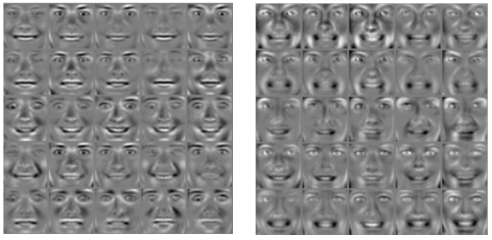}
\caption{Two sample blocks of components of a tri-linear model from \cite{Desjardins2012DisentanglingEntangling}. 
}
\label{fig:bilinear_example_2}
\end{figure}

\subsubsubsection{Trilinear Models}
An example of a \emph{trilinear} model, with 3-way multiplicative interactions, is shown in 
Figure~\ref{fig:bilinear_example_2}. This example comes from 
\cite{Desjardins2012DisentanglingEntangling}. 
The model is trained on the Toronto Face Dataset, composed
of grayscale 48 $\times$ 48 images. 
The two major factors of variation in this dataset are identity
and emotion. The authors also model a third factor, which I will call "group". The third factor
allows the model to represent local consistency without requiring global consistency. Identity
and expression are consistent within a group, but are not necessarily consistent outside of the group.
This makes the modeling easier: we do not have to enforce global consistency of factors.
The groups work on $5 \times 5$ blocks of bilinear components.
In this case, the model is trained in an unsupervised fashion.

The factors themselves are what is known
as "spike-and-slab" variables: each variable has a binary "spike" sub-variable, indicating whether the
variable is on, and a continuous "slab" sub-variable, indicating the weight of the factor. 
A the coefficient for a component is only nonzero when all three "spike" variables are 1; in this way, the value of a
factor acts as a "gate" on the other factors.
In each block, the rows represent identity and the columns represent emotion. Additionally, the blocks
provide local consistency: the identity in the first row of the left-hand block does
not correspond to the identity of the first row of the right-hand block.
A similar \textsc{Multilinear} model is used in a supervised setting \cite{Reed2014LearningInteraction} to separate face identity,
emotion, and pose factors.

\textsc{Multilinear} models are a good choice for environments where a few known factors are known
to exist, and are known to have strong interactions.
In terms of inductive bias, \textsc{Multilinear} models are weaker than \textsc{ISA}.
They allow for multiplicative interactions between coefficients, whereas \textsc{ISA} only
allows linear combinations. This is reflected in the model chart in Figure~\ref{fig:model_chart}.
A sparse bilinear model can be reduced to \textsc{ISA} by setting the all weights of one
factor pool to 1. It can be reduced to \textsc{ICA} by reducing the model to one factor.

Next I will describe a type of environment where factors represent non-overlapping parts of
the observation.

\subsubsection{Functional Parts}
\gen{Generative Assumptions.}\quad 
In some environments, each factor is responsible for generating only a part of
a given observation. For example, a face can be broken down into a number of independent 
parts: eyes, eyebrows, mouth, nose, hair, etc.
There is no overlap between any two parts, so there are no interactions between the factors.
Models suited to these environments are often called "parts-based" or 
"functional parts" \cite{tenenbaum1994functional} models.
In a linear environment, factor components are combined through summation to construct the output.
In a functional parts environment, factor components are \emph{concatenated} to construct the output.

I will describe two methods for learning functional parts models: first, by restricting
coefficients on factors be only be positive, and second, by imposing competition between
factors to explain parts of the input observation.

\subsubsubsection{Non-Negative Matrix Factorization (NMF)}
\ind{Inductive Bias.}\quad 
One model capable of learning parts-based representations is non-negative matrix factorization
(\textsc{NMF}) \cite{Lee1999LearningFactorization.}. Like \textsc{PCA}, \textsc{NMF} reconstructs
an observation by multiplying a weight matrix with a representation vector. However,
both the weight matrix and representation vector are constrained to be non-negative,
which corresponds to the following loss function:
\begin{equation}\label{eq:nmf_loss}
    \mathcal{L}_{\mathrm{NMF}} = \lVert \mathbf{x - \mathrm{zW^{-1}}} \rVert_2^2 \quad s.t.~\mathbf{W}^{-1} \ge 0,~\mathbf{z} \ge 0
\end{equation}
In this case, only positive coefficients, and positive component vectors are allowed. 
In \textsc{PCA}, we can generate a zero output for one of the observed dimensions by 
summing a positive component with a negative component, which cancel each other out.
In \textsc{NMF}, this cooperation is not allowed: the only way to get a zero output is
for both components to be zero. \textsc{NMF} eliminates a type of cooperation between
factors, which encourages them to represent independent portions of the input.

This corresponds to the generative model --- a functional-part factor cannot remove 
anything from another functional part; it can only
contribute positively to the observation. This bias leads to highly localized components.

\rep{Representation Structure.}\quad 
The structure of representations learned by \textsc{NMF} is the same as \textsc{ICA} --- each factor's coefficient
is one scalar value.

\begin{figure}[h]
\centering
\includegraphics[scale=0.3]{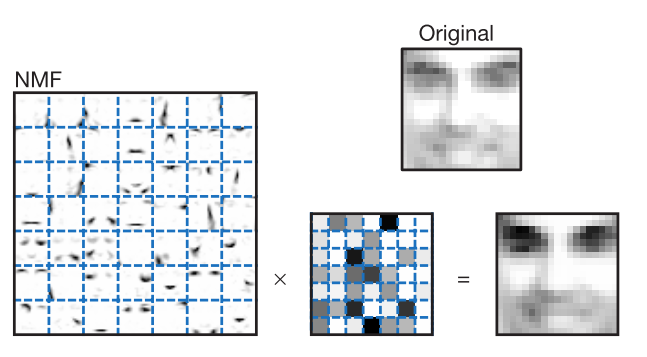}
\includegraphics[scale=0.3]{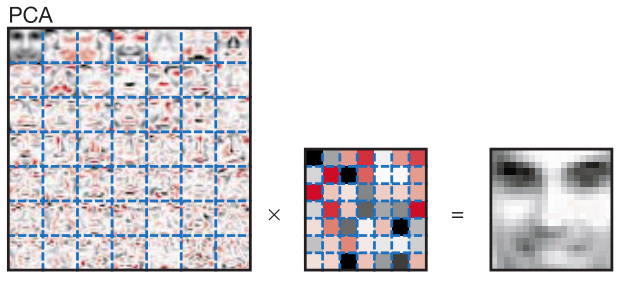}
\caption{}
\label{fig:nmf_example}
\caption{Visualization of a face representation learned by non-negative matrix factorization (\textsc{NMF}), compared
to \textsc{PCA}. Positive component pixels are shown as dark regions, zero pixels are shown as white, and negative
pixels shown as red. Note that the \textsc{NMF} representation has no negative components.}
\end{figure}

\examp{Examples.}\quad 
Figure~\ref{fig:nmf_example} shows two example parts-based representations of faces from
\cite{Lee1999LearningFactorization.}. The representations were learned on a database of 19 $\times$ 19
face images. A visualization of the components (again, the rows of
$\mathrm{W^{-1}}$) for each model are shown. Also shown are the $\mathrm{z}$ weights and reconstruction
for an example face, marked "Original". While the reconstructions are about equally good for
both models,
the \textsc{NMF} representation is clearly parts-based --- some factors correspond to eyes, noses, or
a mouth. In \textsc{PCA}, as before, the factors are highly global.

The inductive bias of \textsc{NMF} is fairly high. It is still essentially linear, with the exception of
the non-negativity constraint. It obviously has higher bias than \textsc{PCA}, since it is a subset of \textsc{PCA},
but it is unclear whether its bias is higher than \textsc{ICA}. It therefore resides at the same coordinate
as \textsc{ICA} in the model chart in Figure~\ref{fig:model_chart}.

\subsubsubsection{Other Models with Functional-Parts Bias}
\textsc{NMF} is not the only means for discovering parts-based representations.
In \cite{Tenenbaum1995FactorialFeatures}, a factorial parts-based model is learned by restricting
each element of the observation to correspond to only one explanatory factor of a vector quantizer.
This can be viewed as another application of the notion of sparsity:
Instead of sparse factor coefficients as we saw in \textsc{ICA}, sparsity is applied to the 
incoming weights from the factors to the observed feature, so that each feature in an
observation is generated by only one factor. This basic approach is also used in
\cite{Ross2003MultipleQuantization}. 

This "functional parts" bias can easily lead to non-factorial representations if used in improper environments. 
In the face example, the face parts are, of course, not independent. The morphology of the face is 
highly influenced by gender and ethnicity, and of course correlations due to symmetry. However,
there are tasks where we want to ignore these types of correlations
and treat these parts as independent anyway.

Next, I will explore a type of bias that assumes that observations are generated by a sequence
of decreasingly abstract layers.

\subsubsection{Hierarchical Layers}
\gen{Generative Assumptions.}\quad 
Some environments have multiple levels of abstraction. For example, a nature scene might
be composed of a foreground and background. First, we choose the locations of the foreground
and background. The foreground is composed of a group of trees and rocks, which are sampled
independently. We then choose a shape of each rock, which is independent of the shape of the other rocks.
In this way, each layer represents a set of choices that are conditionally independent of each other
given the higher level layer.

\subsubsubsection{Recursive ICA (R-ICA)}
\ind{Inductive Bias.}\quad 
Recursive ICA (or \textsc{R-ICA}) \cite{Shan2006RecursiveICA} is a multi-layered version of \textsc{ICA},
where the sparse \textsc{ICA} algorithm is applied iteratively until the desired number of
layers is reached. The first representation layer of \textsc{R-ICA}, $\mathbf{z^1}$, is just the output of \textsc{ICA}. 
The first layer is then prepared for the next layer of \textsc{ICA} by
transforming the outputs $\mathbf{z^1}$ by a function $t(\cdot)$. The purpose of $t(\cdot)$ is to
reformat the highly-non-Gaussian coefficients of $\mathbf{z^1}$ to be Gaussian.
\textsc{ICA} is then run on the Gaussianized first layer to form the second layer $\mathbf{z^2}$: 
\begin{equation}
\begin{split}
    \mathbf{z^1} & = \mathbf{ICA}[~\mathbf{x}~] \\
    \mathbf{z^2} & = \mathbf{ICA}[~t(\mathbf{z^1})~]
\end{split}
\end{equation}
This procedure is repeated until the desired number of layers is reached.
The details of the Gaussian transformation procedure can be found in 
\cite{Shan2006RecursiveICA}, and involves discarding
the sign of the representation, transforming it to a uniform distribution by applying the CDF of the prior,
and then transforming that result back to a Gaussian by using the inverse CDF of the Gaussian distribution.
The signs are discarded because they do not carry any redundancy in the signal, and removing them
helps to reveal nonlinear dependencies.

\rep{Representation Structure.}\quad  The structure of \textsc{R-ICA} is multi-layered, and can be
extended to an arbitrary number of layers.

\begin{figure}
\centering
\begin{subfigure}[b]{0.3\textwidth}
\centering
\includegraphics[scale=0.45]{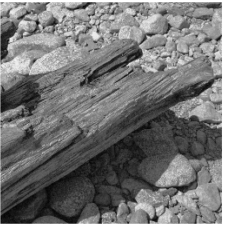}
\end{subfigure}
~
\begin{subfigure}[b]{0.6\textwidth}
\centering
    \begin{tikzpicture}
        \node[anchor=south west,inner sep=0] at (0,0) {\includegraphics[scale=0.25]{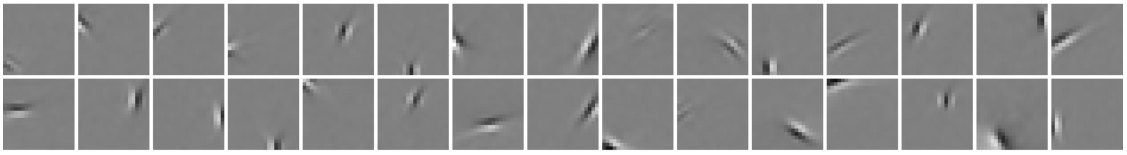}};
        \draw[red,thick,rounded corners] (0,0) rectangle (10, 1.4);
    \end{tikzpicture}
    \begin{tikzpicture}
        \node[anchor=south west,inner sep=0] at (0,0) {\includegraphics[scale=0.25]{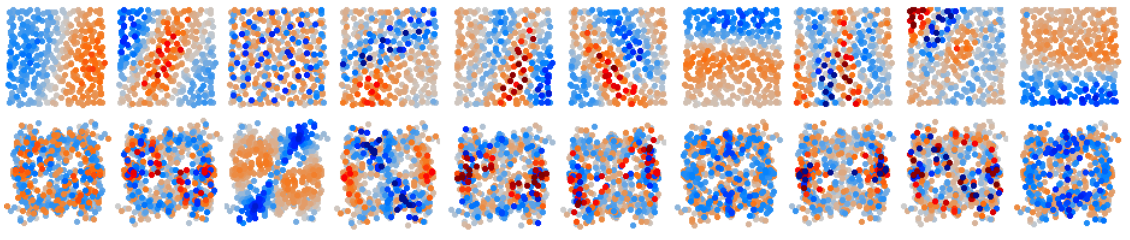}};
        \draw[red,thick,rounded corners] (0,0.1) rectangle (10, 1.05);
        \draw[red,thick,rounded corners] (0,1.15) rectangle (10, 2.1);
    \end{tikzpicture}
\end{subfigure}
\caption{(left) A sample from the dataset of naturalistic images taken from \cite{Shan2006RecursiveICA}.
(top right) Sample  components from the first layer of an \textsc{ICA} solution, trained on 20 $\times$ 20 image patches. 
Each component is a gabor-like filter, with dark regions representing negative pixels and light regions positive.
(center right) Weights from the first layer of \textsc{ICA} to a selection of the second-layer factors. Each square represent
the weights from all first-layer factors to one second-layer factor.  Each dot in a square represents the weight of one
of the components of the first layer, where warmer colors represent a positive weight and cooler colors negative. 
The dots are arranged based on the location of the center of the filter.
(bottom right) Factor coefficients from the second layer of \textsc{ICA}. The dots in this panel arranged by the frequency
and orientation of the fitted Gabor filter, in polar coordinates. 
}
\label{fig:rica_example_1}
\end{figure}
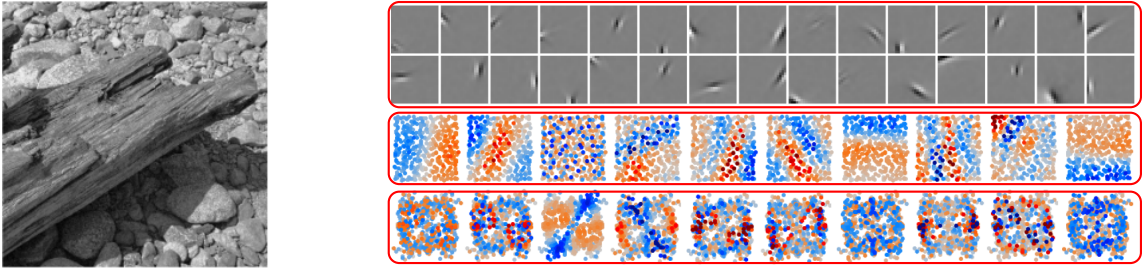

\examp{Examples.}\quad 
An example of \textsc{R-ICA} is shown in Figure~\ref{fig:rica_example_1}. The algorithm was trained on a dataset
of naturalistic images. A sample of Gabor-like \textsc{ICA} components is shown, along with responses from the
second layer of \textsc{ICA}, arranged by both the position of the Gabor-like filter or the frequency and orientation.
Some second-layer factors are strongly position-oriented, whereas others seem to be oriented towards patterns
of frequency and orientation.

\subsubsubsection{Other Models with Hierarchical Layers Bias}
A similar approach for learning higher-order statistical regularities using a hierarchical \textsc{ICA} framework is 
explored in \cite{karklin2003learning}. \textsc{ISA} can also be made into a hierarchy \cite{Le2011LearningAnalysis}, 
to form hierarchical layers of invariant factors.
An \textsc{R-ICA} representation is less restrictive than \textsc{PCA} and \textsc{ICA}; it enables the discovery of more
abstract factors. \textsc{ISA} is also a multi-layered method, and in fact we can construct \textsc{ISA}-like representations
with an \textsc{R-ICA} model. Since \textsc{R-ICA} is a superset that includes \textsc{ISA}, it has lower inductive bias than \textsc{ISA}.
However, \textsc{R-ICA} does not model any multiplicative interactions, we cannot say whether it has lower or
higher inductive bias than a \textsc{Multilinear} model, and therefore appears in the same location on
the chart in Figure~\ref{fig:model_chart}.
A layered or "deep" mixture of experts model \cite{Eigen2014LearningExperts} was found empirically to represent the
location of an object in the first layer of experts, and the identity of an object in the second layer.

These models are all trained sequentially; the first layer is estimated, then its output is used as input
into the next layer. 
A neural network is a variant of a hierarchical model in which the entire model is trained simultaneously.

\

I have so far described three different unsupervised inductive biases for factorial 
representation learning:
distributional bias, invariance bias, and combination bias.
I will now describe the another type of inductive bias, the use of a supervisory signal, which 
can be combined arbitrarily with the last three.


\subsection{Supervision Bias}
Supervisory data are another source of bias, in which we use additional
information connected to observations to constrain the representation of a factor.
For an identity classification task on a face dataset, the 
face identity would be a very useful factor to represent explicitly, and in a way that is not 
confounded with any other non-identity factor. 

A supervised signal can make representation learning completely unnecessary: the
representation is already available. However, the supervisory signal might not always
be available, or it might be weak. We may need a model that works for new data that
are not yet labeled. The supervisory signal may specify only one of many factors in the
representation, and the others still need to be learned. Supervisory signal might be
available for only a portion of the training data. Supervisory signals can also
come in weaker forms that provide constraints on the representation, but do not
specify it directly.

Supervised factors help make representations more interpretable,
to the extent that the factors described by the supervisory data are interpretable.
Supervision is also useful for learning factorial representations in the case of
a non-factorial dataset (sampling bias) or environment.

Supervised constraints come in several different types. Which type to use depends on
both the availability of data as well as on the type of representation sought.

\subsubsection{Constraint Types}
Various supervisory constraints are listed in Table~\ref{tab:constraint_types}.  
They are sorted by how strongly they constrain the solution, with the strongest constraints
at the top. Specifically, I consider how big the constrained space of possible learned solutions is, 
given the value for one of the observations. In this table, $n$ represents the dimensionality
of the representation, and $z(a),z(b),z(c),z(d)$ stand in for representations of different observations
$x(a), x(b), x(c), x(d)$.

\begin{table}[ht]
    \centering
    \caption{Types of constraints.}
    {\rowcolors{2}{gray!10}{gray!0}
    \begin{tabular}{||c|c|c|c||}
    \hline
    \textbf{Type} & \textbf{Example Constraint} & \textbf{Size of constrained space, given a}\\ \hline
    direct & $\mathrm{z(a) = \phi}$ & - \\
    equality & $\mathrm{ \lVert z(a) - z(b) \rVert = 0 }$ & 1 \\
    distance & $\mathrm{\lVert z(a) - z(b) \rVert = \delta}$ & $S(\mathrm{z(a), \delta})$ \\
    inequality & $\mathrm{ \lVert z(a) - z(c) \rVert < \lVert z(a) - z(b) \rVert }$ & $ \mathbb{R}^n + B(\mathrm{z(a), \mathbb{R}^+})$ \\
    analogy & $\mathrm{ z(a) - z(b) = z(c) - z(d)}$ & $\mathbb{R}^{2n} $ \\
    \hline
    \end{tabular}
    }
    \label{tab:constraint_types}
\end{table}

For each type of constraint, I will go into some detail into the types of tasks and representations
that the constraint is well-matched to. I will also discuss the strength of the constraint, and the 
reasoning for its position in Table~\ref{tab:constraint_types}. Then I will present examples from
the literature that illustrate the incorporation of these constraints into models.

\subsubsubsection{Direct}
Direct constraints are most commonly used for attributes of data that we can easily assign
a value to --- a binary attribute, categorical label, or some scalar quantity. For example,
the class of an image of a handwritten digit is a categorical variable of a closed set, zero
through nine.
Direct constraints are the strongest constraints
considered here, and explicitly state the setting for a factor.

\subsubsubsection{Equality}
The most basic non-direct constraint is an \emph{equality}, which simply groups examples together that
are equal with regard to a particular factor.  Equality constraints can be simpler to generate than
direct constraints --- for example, we can ask humans if two face images are the same person or different
identities, rather than asking them to label identity for each image in the dataset.

Equality constraints are useful for learning open-class representations of a categorical 
variable. For example, a face database might have an identity label associated with each
image. If the task is to learn a representation useful for face verification of subjects that are not in
the training database, it is not so useful to simply
learn a categorical representation indicating which of the N people are represented in an image. Instead,
we'd prefer a representation that could accommodate new identities. Instead of directly constraining
the representation, we can constrain the representation to satisfy the equalities in the database.

An equality constraint is simply the specification that two 
examples are equal. This is a strong constraint, since given one observation, the other is constrained 
to occupy exactly the same point.

\subsubsubsection{Distance}
Another common constraint type is a \emph{distance}, or, equivalently, a \emph{similarity} constraint. 
Distance constraints are commonly used in one of three ways: to quantify
some scalar difference between two objective factor values (e.g. difference in age between people), 
to specify an objective similarity between two objects (e.g. $L_2$ vector distance), 
or to specify a human's subjective similarity rating between two objects.

Objective similarities can come from differences between very high dimensional feature vectors.
These are notoriously difficult to visualize, and so a representation that respects these distances 
is useful. For example, images tend to be very high dimensional, and it is difficult
to visualize the relationships between many images simultaneously. 
Another common source of distance information
are object co-occurrences --- for example, we can rate word similarity based on how often they occur
together in a sentence.

Subjective similarities come from humans.
For example, we might want to create a visualization of a dataset of birds, designed to help
us understand human judgments of bird similarity. This visualization could be a Cartesian
space where each point represents a bird, and distance between two birds is reflective of
human judgments of similarity. This visualization could help us understand what
features humans attend to when making similarity judgments. 
In this case, a direct constraint would be a specific coordinate 
value for each bird. However, human judgment of image similarity is known to be complex, and is based on
the simultaneous comparison of many features. Therefore, this labeling task is too difficult for humans. 
In this setting, a direct or equality constraint is not appropriate.

We could collect human judgments of similarity of pairs
of bird pictures, and use these judgments to constrain our representation.
This is the motivation for solutions like multidimensional scaling (MDS) \cite{kruskal1964multidimensional}. 
MDS creates a Cartesian space in which distance between the pair is 
reflective of similarity ratings from humans. 
Each pair of objects with a distance constrains the solution, without 
specifying exactly the coordinates for the pictures. 

Many models dealing with distances are designed to learn one space that satisfies all constraints. 
However, this one-space assumption is not valid in the case of multiple factors which might contribute to a comparison. 
For example, consider the case of \emph{intransitive similarities}.
For example, a red bird is similar to a red fish with respect to color. The red bird is also similar to
a blue bird with respect to animal type. However, the red fish is not similar to a blue bird.
If we correctly factorize the space, such that distance constraints are assigned to color and animal type, 
it is easy to satisfy these constraints.

A \emph{distance} constraint is almost as strong as an equality
constraint.  Given two observations $\mathrm{x(a),~x(b)}$ and a distance between them $\delta$,
the constraint on the representation is $\mathrm{ \lVert z(a) - z(b) \rVert = \delta}$.
Given a representation of the first observation $\mathrm{z(a)}$, the other point $\mathrm{z(b)}$ is constrained
to lie on the sphere around $\mathrm{z(a)}$, with radius equal to the distance $\delta$, or $S(z(a),\delta)$. This is reflected
in Table~\ref{tab:constraint_types}. 

\subsubsubsection{Inequality}
Human judgments of distances
are known to be unreliable --- different users will employ different scales to judge objects, and 
sequential effects can cause drift in judgments based on the order of presentation of stimuli. 

A solution to these issues is to collect an \emph{inequality} constraint, instead of a distance.
Inequality constraints are commonly used to collect human judgments of similarity, while avoiding
the issues with direct similarity judgments.
To acquire an inequality constraint, we present the human labeler with an anchor object $\mathrm{x(a)}$, and 
ask him which of two alternatives,  $\mathrm{x(b)}$ or $\mathrm{x(c)}$, is most similar to the anchor object. 
If the user chooses $\mathrm{x(c)}$, then the resulting 
inequality is: $\mathrm{ sim( x(a), x(c) ) > sim( x(a), x(b) ) }$. We can use this
as a constraint on the representation $\mathrm{z}$ to encourage distance to represent (dis-)similarity:
$\mathrm{ \lVert z(a) - z(c) \rVert < \lVert z(a) - z(b) \lVert} $. 
The constraint is insensitive to the magnitude of the differences, and is therefore
insensitive to different scales or sequential effects. 

Given one point, an inequality has two degrees of freedom. 
Given the point $\mathrm{z(a)},~\mathrm{z(b)}$ is free to lie anywhere in the space $\mathbb{R}^n$, and 
$\mathrm{z(c)}$ is constrained 
to lie inside the open ball around a with a positive radius in $\mathbb{R}^+$, $B(\mathrm{z(a)},\mathbb{R}^+)$.
Since each choice of $\mathrm{z(b)}$ maps to one open ball constraining $\mathrm{z(c)}$ rather than many, we add
the sizes of the sets $\mathbb{R}^n$ and $B(\mathrm{z(a)},\mathbb{R}^+)$ together.
Since $\mathbb{R}^n > S(\mathrm{z(a)},\delta)$, inequalities are a weaker constraint than distances,
as reflected in its position in Table~\ref{tab:constraint_types}.

\subsubsubsection{Analogy}
Another class of constraints are called \emph{analogies}. Humans are adept at analogical 
reasoning, and are able to extrapolate answers to hypothetical questions such as "What if this face
were viewed from a different angle?" or "What if this piece of music were played at a higher tempo?".

Analogy constraints take the form $\mathrm{x(a) : x(b) :: x(c) : x(d)}$, or, the relationship between observations
$\mathrm{x(a)}$ and $\mathrm{x(b)}$ is the same
as the relationship between $\mathrm{x(c)}$ and $\mathrm{x(d)}$. 
They are a means of capturing a relationship between observations.

Analogy constraints can be convenient in situations where direct constraints are not possible, but
we have access to pairs of observations where we expect their relationship to be the same. For example,
we might have images of two people before and after completing high school, or before and after completing
college. We can interpret two of these pairs as an analogy constraint, and 
use them to learn age as a factor in a representation.

Analogies are typically interpreted as vector-difference constraints on the representation.
For example, "x(a):x(b) :: x(c):x(d)" can be interpreted as $\mathrm{z(a) - z(b) = z(c) - z(d)}$.
Given the representation $\mathrm{z(a)}$, $\mathrm{z(b)}$ is free to lie anywhere in the space $\mathbb{R}^n$.
Given $\mathrm{z(a)}$ and $\mathrm{z(b)}$, $\mathrm{z(c)}$ is still free to lie anywhere else in the space $\mathbb{R}^n$. 
However, given $\mathrm{z(a), z(b)}$, and $\mathrm{z(c)}$, we can
determine the exact location of $\mathrm{z(d)}$, so it does not constrain the space further. So, the size of the constrained set
is $\mathbb{R}^{2n}$. Since $\mathbb{R}^{2n} > \mathbb{R}^n$, we can see that analogies are a weaker constraint than 
inequalities, and are the lowest level of supervision in Table~\ref{tab:constraint_types}.

\subsubsubsection{Summary of Constraint Types}
Direct constraints are useful for learning representations where factors
have an unambiguous category or quantity.
Equality constraints are almost as strong, are
easier to obtain, and can be useful for learning open-class representations of a categorical
variable. Objective measures of similarity are best modeled as distances,
and human judgments of similarity are best modeled as inequalities. Analogies are the
weakest constraint, and are applicable in situations where only the relationships 
among observations are available.

The best representation for a task might not
correspond to the type of constraint available. For example, we might
have direct constraints for face identity in the form of categorical labels. 
However, we might not want to represent faces with a categorical variable.
Instead, we might choose a representation where distance is related to the 
similarity of faces, so that we can understand which faces are most similar.
In this case, the direct constraints can be transformed into equality constraints.

I will present models showing examples across the entire spectrum of constraint types.
Each model shown will present some evidence for factorial representations.
Before presenting these examples, I need to discuss how I will evaluate the factorial representations in the papers.

\subsubsection{Evaluation of Representations}
In a supervised context, we have the opportunity to directly measure some properties
of factorial representations.
There are three broad categories of evidence that demonstrate factorial representations:
\begin{itemize}
    \item \textbf{Distillation.} Information relating to a supervised factor should be 
    distilled into the factor representation.
    This distillation should make the information more easily accessible, as evidenced
    through results such as improved classification performance, or example reconstructions
    that traverse the factor manifold showing variation in that factor.
    \item \textbf{Disentangling.} Information not relating to a supervised factor
    should no longer be present in the factor's representation. We should be able to see
    a decrease in performance on a \emph{non-target} task using the representation of
    a factor. We should also find evidence of \emph{invariance} to other factors of variation.
    \item \textbf{Cross-Over Effects.} In the case of partially-labeled factors, 
    disentangling should have the effect of reducing
    nuisance variation on even the non-supervised factors. We should therefore
    expect to see better factor distillation in even non-supervised factors, as
    evidenced by improved classification performance, or example reconstructions.
\end{itemize}

Next, I will describe example models where constraints are used
to learn factorial representations. For each model, I will summarize the constraints,
the method used for incorporating the constraint, and show examples that demonstrate evidence
of factorial representations, where available.
I also describe some modes that combine several different types of constraints.

\subsubsection{Examples of Models}

\subsubsubsection{Semi-Supervised VAE (Direct)}
Direct constraints are most commonly applied by directly encouraging the representation
to mirror the labels.
For example, in \cite{Kingma2014Semi-supervisedModels},
a variational auto-encoder (VAE) is modified to also include a categorical label for each data 
point, to form a Semi-Supervised VAE (\textsc{SSVAE}).

In the standard VAE \cite{Kingma2013Auto-EncodingBayes} observations are generated 
by first choosing a representation $\mathrm{z}$, 
from a zero-mean unit Gaussian distribution with diagonal covariance --- corresponding to
a bias of independent Gaussian factors. The observation is then
generated through a non-linear function of the representation (a neural net), 
with parameters $\theta$.
\begin{equation}
p(\mathbf{z}) = \mathcal{N}(\mathbf{z}|0, \mathbf{I}) \qquad
p_\theta(\mathbf{x}|\mathbf{z}) = f(\mathbf{x}; \mathbf{z, \theta})
\end{equation}

In the \textsc{SSVAE} \cite{Kingma2014Semi-supervisedModels}, we first generate a categorical label
from the multinomial distribution $\mathrm{Cat(y|\pi)}$, with parameters $\pi$.
We independently sample a $\mathbf{z}$. The observed data $\mathbf{x}$ are then
generated conditionally on both $y$ and $\mathbf{z}$:
\begin{equation}
p(y) = \mathrm{Cat}(y|\mathbf{\pi}) \qquad 
p(\mathbf{z}) = \mathcal{N}(\mathbf{z}|0, \mathbf{I}) \qquad
p_\theta(\mathbf{x}|y,\mathbf{z}) = f(\mathbf{x}; y, \mathbf{z, \theta})
\end{equation}

The \textsc{SSVAE} can account for \emph{partially-labeled} datasets by marginalizing over $y$ for
datapoints where no label is available. The representation is then composed of two
factors, $y$, the class label, and $\mathrm{z}$, the remainder of the representation.
The optimization then follows a procedure similar to that of the VAE.

\begin{figure}[ht]
    \centering
    \includegraphics[scale=0.4]{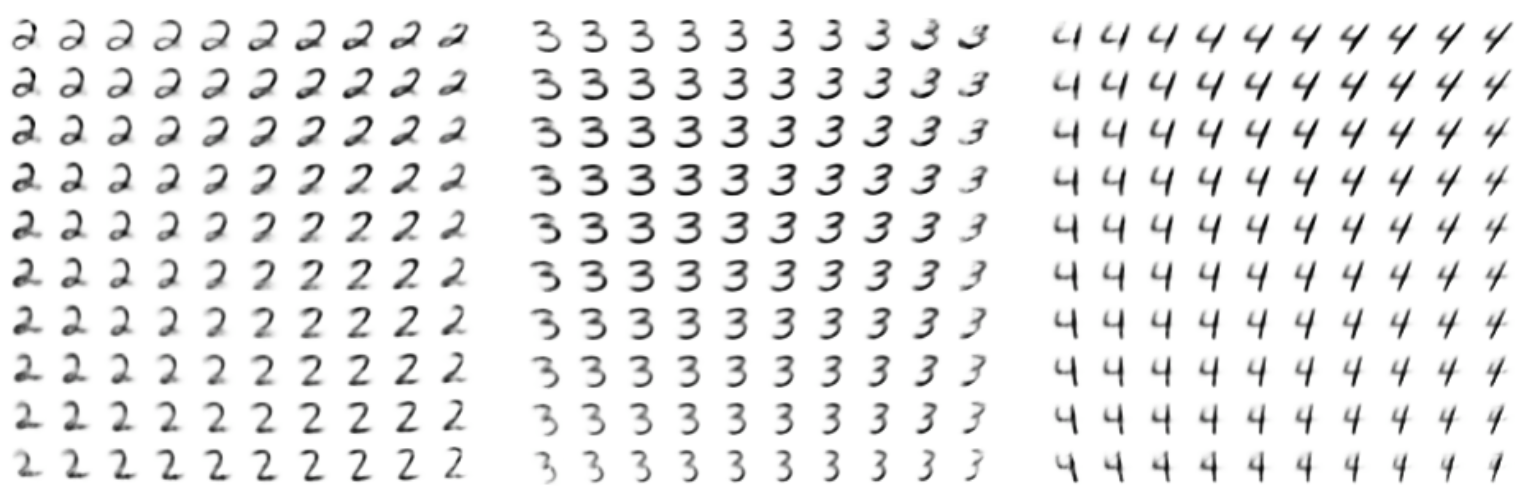}
    \caption{Samples of handwriting styles from an \textsc{SSVAE} trained on the MNIST digit dataset.}
    \label{fig:ssvae_example}
\end{figure}

The authors trained an \textsc{SSVAE} on the standard MNIST dataset, which is composed of small 
black-and-white images of hand-written digits. The authors
show evidence of factor \textbf{distillation}: \textsc{SSVAE} is more accurate than VAE
in classifying the digit identity given an image. 
The authors also demonstrate \textbf{cross-over effects}: 
by representing digit identity using $y$, $\mathrm{z}$ is encouraged
to represent \emph{other} sources of variation. Figure~\ref{fig:ssvae_example}
shows an example of this effect. The model is trained using a two-dimensional $\mathrm{z}$
vector. Each image block are samples from the model with $y$ clamped to represent the
digit 2, 3, and 4. Within each block, the dimensions of $\mathrm{z}$ are varied smoothly.
Nearby samples correspond to similar \emph{styles} of handwriting: the left region
corresponds to non-slanted styles, and the right region represents slanted styles.

\noindent \textbf{Similar Models.}
Many other semi-supervised models use direct constraints in this fashion. For example,
in \cite{rasmus2015semi},
the authors add class labels to part of the representation of a ladder network, a form
of deep denoising autoencoder. They report state-of-the-art performance on
semi-supervised MNIST and CIFAR-10 tasks.
There are a few examples of using direct constraints alongside with higher encoder inductive biases. For example,
there are supervised versions of \textsc{ICA} \cite{sakaguchi2002feature}, and \textsc{NMF} \cite{wang2009non}. 

Direct constraints are the most straightforward constraints, and are appropriate for types
of factors that are easily quantified, such as binary, categorical, or numerical variables.
Next I will describe a weaker kind of constraint, where the exact value of a factor is not known.

\subsubsubsection{Inverse Graphics Network (Equality)}
The Deep Convolutional Inverse Graphics Network (\textsc{DC-IGN}) \cite{Kulkarni2015DeepNetwork} uses
equality constraints on a set of synthetically-generated face images. 
An autoencoder is trained to 
explicitly represent a number of factors, including azimuth angle of the face, elevation angle,
and azimuth angle of the light source. The faces are synthetically generated, so the authors have
a unique ability to put constraints on \emph{all} of the factors of variation in the images. 

The \textsc{DC-IGN} representation is split up to represent three factors explicitly, and to represent
the remaining factors using a single pool of neurons.
The first factor neuron represents the face azimuth, the second the face elevation, and the 
third represents the light source azimuth.
The remaining pool of neurons represents other nuisance factors, including face identity, morphology, and expression.

The authors developed a custom auto-encoder training procedure that varies one factor at a time while holding
all other factors of variation constant. The representations of the non-varying factors are encouraged by the
learning procedure to remain constant. In the training, the pool representing nuisance factors is treated
as one factor.

\begin{figure}[ht]
    \centering
    \includegraphics[scale=0.3]{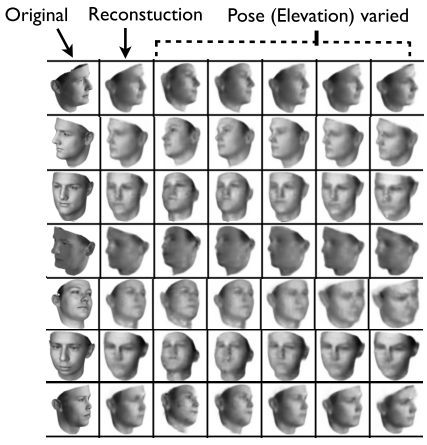}
    \qquad
    \includegraphics[scale=0.5]{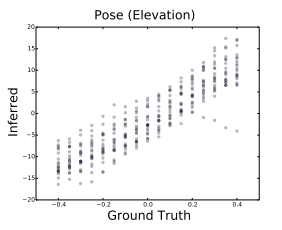}
    \caption{(left) Varying the elevation of random test samples.
    (right) A scatter plot showing the inferred elevation factor representation vs. the
    ground truth setting used to render the face image.}
    \label{fig:ign_example}
\end{figure}

Figure~\ref{fig:ign_example} shows some sample reconstructions from the network, which shows
evidence for factor \textbf{distillation} and \textbf{disentangling}. On the left, we see
some sample input images and reconstructions. The remainder of the grid is formed by manipulating
the representation of elevation, changing it smoothly over a range of values, and leaving the other factors
unchanged. We can see in the reconstructions that face elevation is changing, but the face azimuth,
light source, and identity remain constant. The authors demonstrate variation in the same way on
the other factors as well. The authors did not explore cross-over effects.

On the right, the ground truth elevation of a sample of synthetic images is plotted
against the learned representation of elevation. We can see that the two variables are highly
correlated.

\subsubsubsection{Siamese Network (Equality)}
As noted, equality constraints are useful for learning open-set representations of categorical
variables. This procedure is also called a \emph{learned similarity metric}.
The method is shown off in the \textsc{Siamese} network of \cite{Hadsell2006DimensionalityMapping}, which
learns a low-dimensional embedding of MNIST digits where similarity corresponds to likelihood of being
in the same digit class. \cite{Chopra2005LearningVerification} used a \textsc{Siamese} network to build an embedding
representation of face identity. Instead of the clamping procedure, \textsc{Siamese} networks are trained with a
loss which minimizes distance for same-category examples and maximizes distance for out-of-category
examples. Given two examples $\mathbf{x}\mathrm{(a)}, \mathbf{x}\mathrm{(b)}$, 
whose factor values of factor $\mathrm{f}$ are $\mathrm{f(a)}$ and $\mathrm{f(b)}$,
and their corresponding representations $\mathbf{z}^\mathrm{f}\mathrm{(a)}, \mathbf{z}^\mathrm{f}\mathrm{(b)}$,
the loss function in \cite{Hadsell2006DimensionalityMapping} is:
\begin{equation}\label{eq:siamese}
    \mathcal{L}_\mathrm{SIAMESE} = 
    \begin{cases}
    \lVert \mathbf{z}^\mathrm{f}\mathrm{(a)} - \mathbf{z}^\mathrm{f}\mathrm{(b)} \lVert_2^2 & \text{ if } \mathrm{f(a)} = \mathrm{f(b)} \\
    \big[ max(0, m - \lVert \mathbf{z}^\mathrm{f}\mathrm{(a)} - \mathbf{z}^\mathrm{f}\mathrm{(b)} \lVert_2^2 \big] & \text{ otherwise }
    \end{cases}
\end{equation}

In \cite{Zeghidour2016JointNetworks}, this technique is extended to represent multiple
factors simultaneously as separate embedding spaces.
The loss functions for the embeddings are simply summed together
to form the final loss function. 

The authors of \cite{Zeghidour2016JointNetworks} performed experiments using a speech dataset of consisting
of different speakers speaking the same set of words. One embedding was trained to represent
speaker identity, and anther was trained to represent the target phonemes. Instead of the euclidean
distance differences in Equation~\ref{eq:siamese}, they used cosine similarity.
The networks were trained on a subset of the Librispeech dataset, which consisted of 360 hours of read speech from
920 speakers. To test the model, the authors performed an ABX task to estimate classification accuracy.
The error rates reported are the percentage of correct answers to the ABX tests. In the tests, the
examples are triphones (a phoneme in the context of two other phonemes).
To test phonetic discrimination, the A and B examples were from the same speaker, while the A and X
examples matched on phonetic content but were pronounced by different speakers. This allows the
testing to directly measure \textbf{disentanglement} as well as \textbf{distillation}.

\begin{figure}
    \centering
    \includegraphics[scale=0.3]{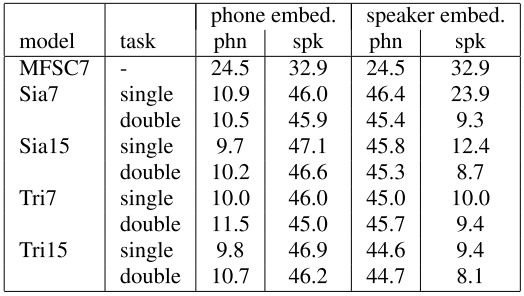}
    \caption{Phonetic and Speaker ABX discrimination results from \cite{Zeghidour2016JointNetworks}. }
    \label{fig:siamese_example_1}
\end{figure}

Figure~\ref{fig:siamese_example_1} shows a table of results from \cite{Zeghidour2016JointNetworks}.
In the table, "Sia" stands for "Siamese" and "Tri" stands for "Triamese", or a network which was trained
using a loss function that took both a positive and negative sample into account simultaneously. The
number after the model name indicates how many stacked frames were included in the input.
"MFSC7" is a baseline model that simply concatenates 7 MFCC frames together and uses that as the embedding.
The "task" refers to whether the speaker and phonetic embeddings were trained separately ("single")
or together ("double"). Performance is reported for both embeddings on both tasks, with lower error
rates indicating better performance.

We can see strong evidence of \textbf{distillation}, as the trained embeddings learn to perform better
at predicting their target factor. We also see evidence of \textbf{disentanglement}, as the trained
embeddings perform worse on the non-target factor. We also see some indirect evidence for 
\textbf{cross-over effects}, since the "double" models tend to perform better than the "single" models,
especially on speaker identification.

\subsubsubsection{Multiple-Maps t-SNE (Distance)}
A model that accounts for distances with multiple factors is \emph{multiple-maps t-SNE} 
(\textsc{mm t-SNE})
\cite{VanDerMaaten2012VisualizingMaps}, 
which is an extension of t-distributed Stochastic Neighbor Embedding (t-SNE) \cite{maaten2008visualizing}.
In t-SNE, distances between pairs of observations $i$ and $j$ are converted into probabilities; they form
a joint distribution P over all pairs of objects where $p_{ij}$ is proportional to the similarity between
objects $i$ and $j$.

The dataset used in \cite{VanDerMaaten2012VisualizingMaps} is composed of human-labeled word associations.
Humans were given a prompt word and were asked to name associated words. 5,019 words were used as prompts,
and 10,617 total words were labeled in associations. Each word and its associations are counted as one
word co-occurrence. Co-occurrences were made symmetric and normalized to probabilities.
In this case, the factors roughly correspond to topics, which can lead to intransitive similarities.
For example, the word "tie" might co-occur frequently with "tuxuedo" when the topic is "formal wear". 
The word "tie" might also co-occur frequently with "rope" when the topic is related to ropes and knots.
However, "rope" will likely not co-occur with "tuxuedo".
A factorial representation should separate these topics and allow the model to represent intransitive similarities.

The goal is to learn a word-representation \textbf{z} in which distances are reflective of the probabilities 
in P. To facilitate this goal, the learning procedure defines a probability distribution Q
over distances in \textbf{z}, and minimizes the KL-divergence $KL(P\lVert Q)$). 

Q is chosen so that distances are proportional to a Student-t 
distribution with one degree of freedom\footnote{The choice of Q is important, because it determines which distances are attended to most during the learning. The t-distribution is chosen instead of a Gaussian because it allows
points that are only slightly similar in high dimensional space to become far apart in the low dimensional
representation, a solution for the crowding problem of high-dimensional spaces.}. The probability $q$ between
words $\mathbf{x}(i)$ and $\mathbf{x}(j)$ with representations $\mathbf{z}(i),~\mathbf{z}(j)$ is defined as:
\begin{equation}
q_{ij} =  \frac{ (1 + \lVert \mathbf{z}(i) - \mathbf{z}(j) \rVert^2 )^{-1} }
{\sum_k{ \sum_{l \neq k}{ ( 1 + \lVert \mathbf{z}(k) - \mathbf{z}(l) \rVert^2 )^{-1} } }}
\end{equation}
The top term means that the probability is inversely proportional to the distance between $\mathbf{z(i)}$ and
$\mathbf{z(j)}$, and the bottom normalization term ensures that $q_{ij}$ behaves like a probability distribution.

This model is extended to include multiple factors.
Each factor defines its own space in which all the words are mapped. For each word $i$ and factor $f$,
the model learns an \emph{importance weight} $\pi^i_f$. Importance weights are between zero and one, and sum
to 1 over all the factors for each word. For a given combination of words $\mathbf{x}_i, \mathbf{x}_j$,
the distance calculation is multiplied by the product of the two factor-specific importance weights. This has the effect
of only penalizing the model for not respecting distances when both words have high importance weight for a factor.
Q is updated to sum over all the factors:
\begin{equation}
q_{ij} = \frac{ \sum_f{ \pi^i_f \pi^j_f ( 1 + \lVert \mathbf{z}_f(i) - \mathbf{z}_f(j) \rVert^2 )^{-1} } }
{ \sum_k{ \sum_{l \neq k}{ \sum_{f'}{ \pi_{(f')}^k \pi_{(f')}^l 
(1 + \lVert \mathbf{z}_{f'}(k) - \mathbf{z}_{f'}(l) \rVert^2 )^{-1} } } } }
\end{equation}

Optimization finds the best settings for $z$ as well as $\pi$ such that the original co-occurrence probabilities in P are respected.
In this way, the model automatically assigns words to factors --- it does not depend on any
prior knowledge of factors or associations between factors and similarity ratings, this makes the constraints
in this model \emph{factor-agnostic}. The number of factors must be known in advance.

\begin{figure}[ht]
    \centering
    \fbox{\includegraphics[scale=0.4,trim={0 4cm 0 4cm}]{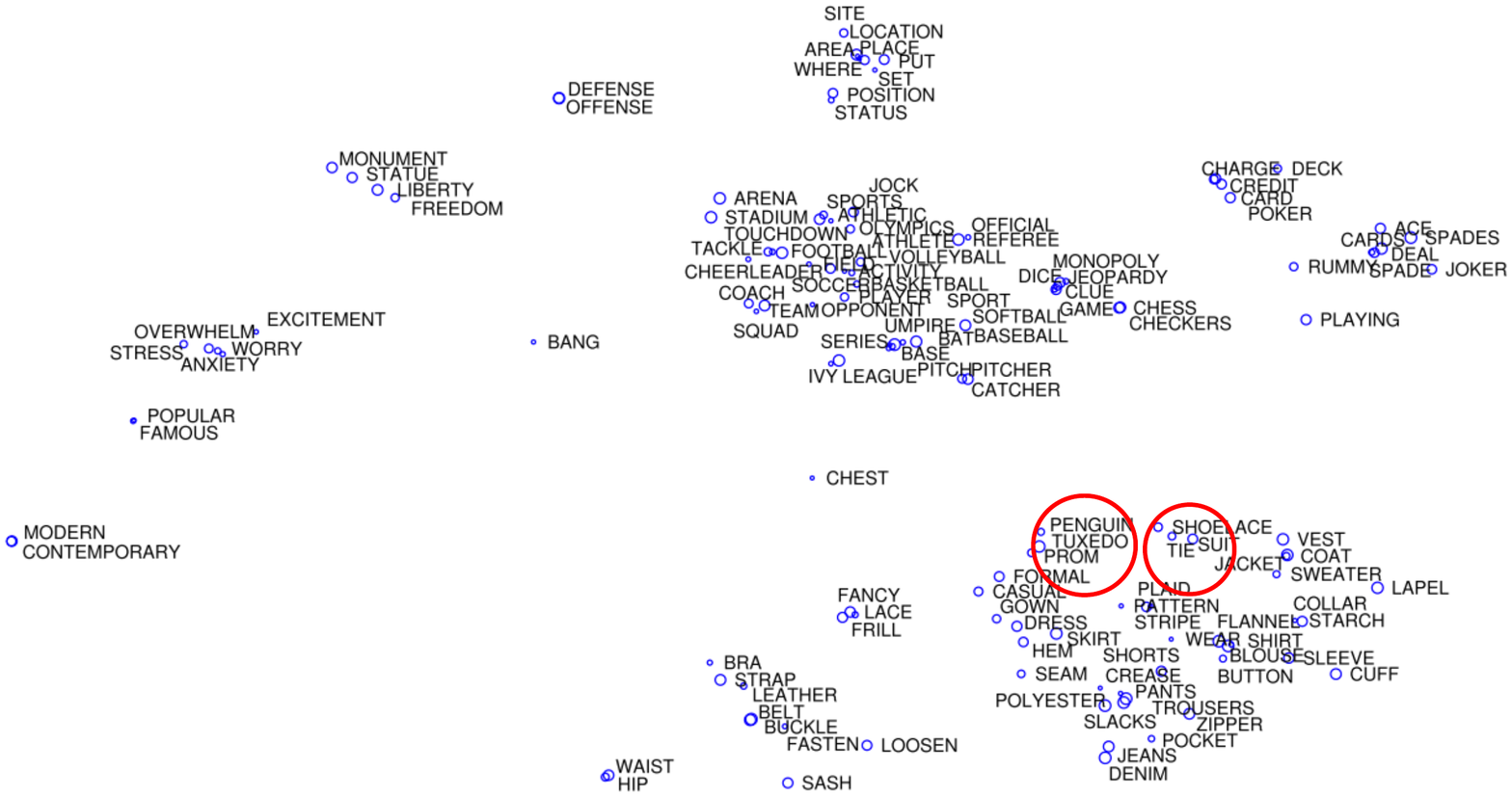}}
    \\
    \fbox{\includegraphics[scale=0.4,trim={0 4cm 0 4cm}]{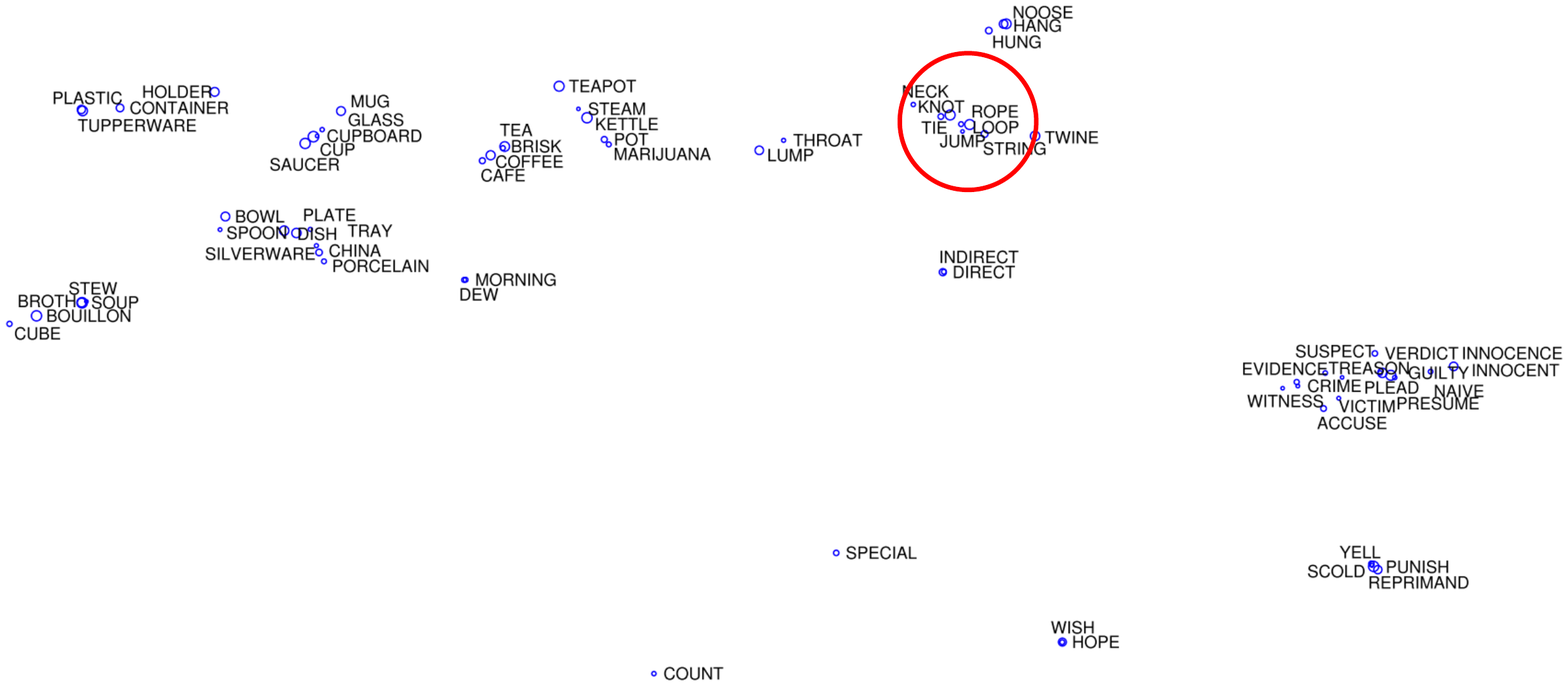}}
    \caption{Visualization of two two-dimensional maps learned via \textsc{mm t-SNE}, trained on
    human-generated word associatation data.}
    \label{fig:mm_t_sne_example}
\end{figure}

Figure~\ref{fig:mm_t_sne_example} shows some example maps from this model. The model was trained with 40
maps/factors, which are all two-dimensional. The dataset contains many topics, so there are typically multiple topics per factor.
While every map contains every word, words with an importance weight below 0.1 are not shown.
The top map corresponds to "sports" and "fashion", whereas the bottom map corresponds to a few topics,
including criminal justice, kitchen equipment, and ropes/knots. We see that "tie" is placed near "tuxuedo"
in the top map, and near "rope" in the bottom map.
Some of the maps have some topic coherency, which is evidence for \textbf{distillation}.
Not every map has every topic in it, which is evidence for
a \textbf{disentangled} representation.

In the case of \textsc{mm t-SNE}, the fact that the constraints are factor-agnostic
severely weakens the constraint space by multiplying the size of the space by the number of possible
factors that a constraint could be assigned to, or $S(\mathrm{z}(a),\delta)$ times the number of factors/maps.

\subsubsubsection{Karaletsos et al. 2015 (Inequality)}
Triplet inequalities are used to learn a factorial representation for a 
variational autoencoder in \cite{Karaletsos2015BayesianConstraints}.
An oracle (which could be human or machine) answers questions in the form of
"is $\mathrm{x(i)}$ more similar to $\mathrm{x(j)}$ or $\mathrm{x(l)}$ in terms
of factor $f$?". The answer to this question is a triplet $t^f_{i,j,l}$,
interpreted as a factor-specific inequality: $\mathrm{ sim_f( x(i), x(j) ) > sim_f( x(i), x(l) ) }$. 
The factors for each inequality are specified in the training signal.
The number of factors and the triplets associated with each factor are specified in advance.

The probability of the inequality is modeled as a softmax function, with a factor-specific
dis-similarity function D whose range only includes values greater than or equal to zero:
\begin{equation}
    p(t^f_{i,j,l}) = Ber(t^f_{i,j,l}) = \frac{ e^{-D^f_{i,j} } }{ e^{-D^f_{i,j}} + e^{-D^f_{i,l}} }
\end{equation}

The model discovers an $N$-dimensional embedding representation, which serves as the basis
for the the $D^f_{a,b}$ function.  In the paper, $D^f_{a,b}$ is defined in terms of the 
Jenson-Shannon divergences between each of the N dimensions of a and b, although in 
practice, an approximation to the true JS divergence is used.

The model learns which of the $N$ dimensions should contribute to explaining each inequality.
A factor-specific \emph{mask} selectively attends to dimensions of the representation that are 
relevant to the factor and hides irrelevant dimensions. 
Each factor-specific mask variable $\mathbf{m^f}$ is $N$-dimensional, and 
constrained to lie in the interval $[0,1]^N$. The distance function $D$ 
takes the mask into account when comparing the dimensions of the representation of two
examples:
\begin{equation}
    \overset{\mathbf{m^f}}{D_{a,b}} = \sum_{n=1}^N \mathbf{m_n^f} D_{a,b}^n
\end{equation}

The optimization involves maximizing $p(t_{i,j,l})$ for all the factor-specific triplets, 
alongside the reconstruction error and representation regularization terms in the VAE.
The masks are learned during the optimization.
In this way, the model learns to switch on an appropriate subspace that represents factor $\mathbf{f}$
to explain the triplets drawn with respect to that factor.

The authors test their model on the Yale Faces dataset, composed of 2,414 images of 38 individuals
under varying lighting conditions. The azimuth and elevation of the light sources are varied in the images.
Each triplet is composed of three randomly selected images.
The three factors of variation --- identity, light source azimuth, and light source elevation --- were
used to create triplet inequalities out of the three randomly chosen images. For example, a light source
azimuth triplet would provide information about which two images had more similar azimuths.

\begin{figure}
    \centering
    \includegraphics[scale=0.4]{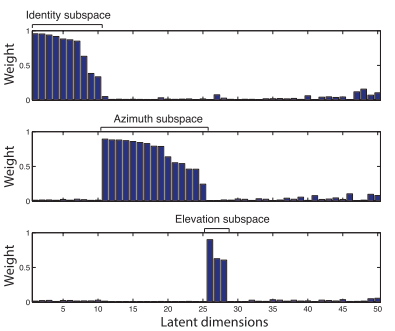}
    \includegraphics[scale=0.4]{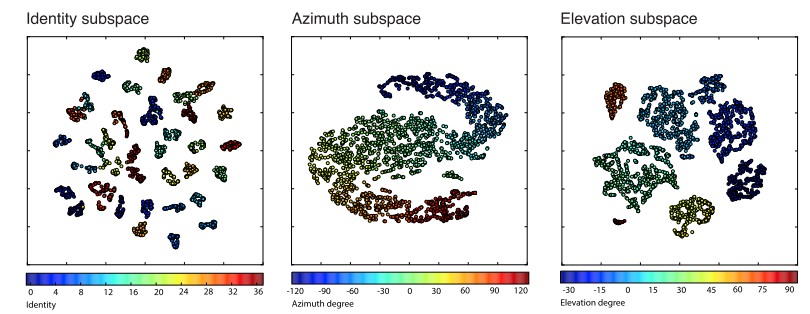}
    \caption{(left) A visualization of three factor-specific masks learned on the Yale Faces dataset, from
    \cite{Karaletsos2015BayesianConstraints}.
    (right) A t-SNE visualization of the identity, azimuth, and elevation subspaces as derived
    from the masked dimensions on the right. Only dimensions with a weight greater than 0.2 were used as
    input for this visualization. Colors represent the actual values for the factor of interest.}
    \label{fig:karaletsos_example}
\end{figure}

The authors demonstrate factor \textbf{distillation}: this masked model performs better on 
identity classification, azimuth prediction,
and elevation prediction, compared to a model trained with no constraints. 
Figure~\ref{fig:karaletsos_example}
shows a visualization of the three factor-specific masks learned on the face dataset. The masks learn to
factorize the space, such that only a subset of the dimensions of z contribute to decisions comparing
identity or light source position. A dimension-reduced visualization of the dimensions in these masked
subspaces is shown on the right in Figure~\ref{fig:karaletsos_example}.  We can see that the subspaces
group similar examples together, as would be expected.

\noindent \textbf{Similar Models.}
A very similar approach is taken in \cite{Veit2016DisentanglingNetworks}, which demonstrated
the model's ability to learn factorial representations on other datasets, such as stylized fonts and images of shoes.
Factor-agnostic triplets are used to generate multi-factor map representations in \cite{Amid2015MultiviewMaps}.
Triplet constraints have also been used to learn highly accurate single-factor embedding models
\cite{schroff2015facenet,wang2014learning}.

\subsubsubsection{Deep Visual Analogy-Making (Analogy and Equality)}
In Deep Visual Analogy-Making (\textsc{DVA}) \cite{Reed2015DeepAnalogy-Making}, 
two different types of constraints are used
to learn a factorial representation in an autoencoder architecture.
\emph{Equality} constraints are used to enforce factorization of the representation
into different attributes and \emph{analogy} constraints are used to shape the
subspace representing each factor. Each of these constraints is incorporated
into an autoencoder architecture via its own constraint-specific loss function.

Assume we have data observations $\mathrm{ x(a), x(b), x(c), x(d) }$
in the analogy $\mathrm{x(a) : x(b) :: x(c) : x(d)}$.
The encoder function $e(\cdot)$ maps an
observation to its representation. The generator function $g(\cdot)$ maps a
representation back to the observational space.
The loss is defined in terms of the 
square difference (in the observation space) of example $d$ with the analogical 
prediction for $d$ given ${a,b,c}$:
\begin{equation}\label{eq:analogy_add}
\mathcal{L}_\mathrm{ANALOGY\_ADD} = \lVert \mathrm{x(d) - g( e[x(b)] - e[x(a)] + e[x(c)] )} \rVert_2^2
\end{equation}
This corresponds to a vector-addition interpretation of the analogy.
However, this interpretation does not make sense for operations such as
rotations. Rotations are circular, so a sufficient rotation should cause
the point to end up back where it started. To address this issue,
the authors include two additional interpretations of the analogy. One
uses multiplicative interactions between $e(b)-e(a)$ and $e(c)$ to
determine the vector that is added to $e(c)$.
The other uses a neural network whose inputs are $e(b)-e(a)$ and $e(c)$.
The neural net perform the best, because it is the most flexible of the three models.

The authors also use an additional loss, like Equation~\ref{eq:analogy_add}, but in
the representation-space rather than the output-space, to enforce the relationships
between the representations.

To learn a factorial representation, the authors use an additional penalty based on
equality constraints.  Suppose we have two non-overlapping sets of factors called
$\mathrm{f1}$ and $\mathrm{f2}$, where $\mathrm{f1} \cap \mathrm{f2} = \emptyset$.
Additionally, all of the factors $\mathrm{f}$ are included: $\mathrm{f1} \cup \mathrm{f2} = \mathrm{f}$.
We choose three example observations a, b, and c,
where a and c share the $\mathrm{f1}$ factors and b and c share the $\mathrm{f2}$ factors.
This corresponds to two factor-specific equalities:
$\mathrm{a}^\mathrm{f1} = \mathrm{c}^\mathrm{f1}$ and 
$\mathrm{b}^\mathrm{f2} = \mathrm{c}^\mathrm{f2}$. 

A binary mask $s$, of the same dimension as the representation, 
serves to select representation dimensions corresponding to $\mathrm{f1}$ vs dimensions
corresponding to $\mathrm{f2}$. The mask $s$ is not learned; it is specified in advance. 
In this case, we expect to be able to reconstruct
c using the dimensions corresponding to $\mathrm{f1}$ from a and the dimensions
corresponding to $\mathrm{f2}$ from b. This corresponds to the following loss function:
\begin{equation}
\mathcal{L}_\mathrm{EQL} = \lVert \mathrm{ x(c) - e( s \cdot e[ x(a) ] + (1 - s) \cdot e[ x(b) ] } \rVert_2^2
\end{equation}

\begin{figure}[h]
    \centering
    \raisebox{5\height}{(a)}
    \raisebox{0.075\height}{\includegraphics[scale=0.35]{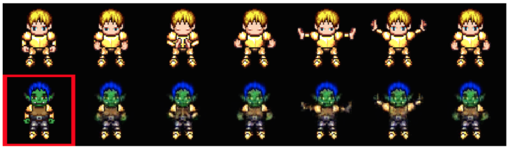}}
    \raisebox{5\height}{(b)}
    \includegraphics[scale=0.25]{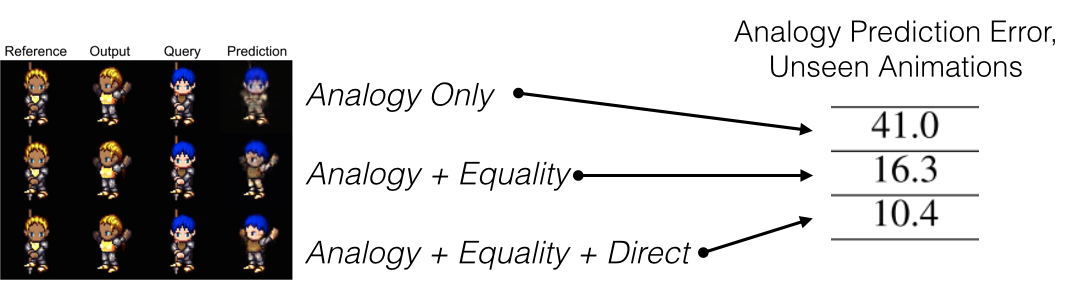}
    \caption{Example predictions from an autoencoder trained using
    analogies, from \cite{Reed2015DeepAnalogy-Making}. 
    (a) "Animation transfer" results from the sprites dataset.
    (b) Analogy prediction results for models trained on three
    different combinations of constraints.
    }
    \label{fig:analogy_example_1}
\end{figure}

To demonstrate the model, the authors train on a dataset of images of video game 
characters, called "sprites", and rotated 3D car renderings.
An example sequence of reconstructions of analogies is shown in Figure~\ref{fig:analogy_example_1} (a),
in which the sequence of pose changes from the character on top is transferred to the character
on the bottom.
Each character has an identity and a pose. Three models were built: the first, trained with only
analogy constraints, the second trained to separate identity from pose using equality constraints,
and the third trained to classify identity with direct constraints and to predict pose with equality
constraints.
According to the constraint hierarchy in Table~\ref{tab:constraint_types}, the model
with the strongest constraints is the third model, followed by the second, then first.
The prediction results, shown in Figure~\ref{fig:analogy_example_1}, from the paper follow this pattern.
The authors demonstrate factor \textbf{distillation} by improved reconstruction performance on test analogies. 


\noindent \textbf{Similar Models.}
Visual analogies are also used to learn representations 
in \cite{JuHwang2013Analogy-preservingCategorization}. Neural net word
representations are known to obey analogical reasoning
\cite{mikolov2013distributed}, but these models are not explicitly trained
using analogy constraints.

\subsubsubsection{Transforming Auto-Encoders (Distance and Equality)}
An approach that combines equality and distance constraints is to teach a model to
\emph{transform} an observation with one setting for a factor, into another observation
with a different known setting for that factor. The discrepancy between the factor values
between the input and output is fed as input into the model. This approach also uses
equality constraints, because it requires that all \emph{other} factors remain equal
while varying the target factor.

\begin{figure}
    \centering
    \includegraphics[scale=0.3]{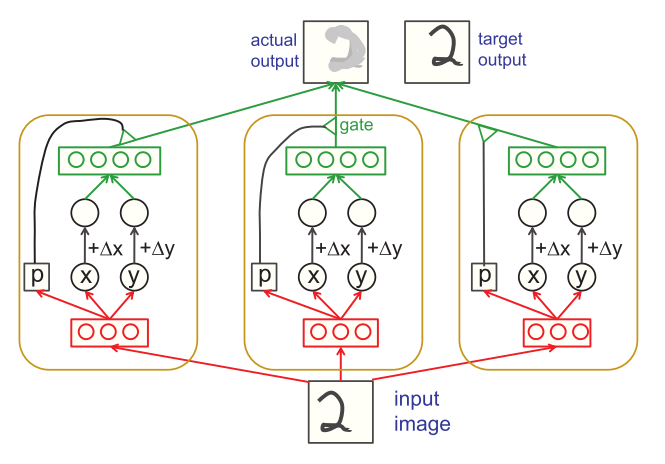}
    \includegraphics[scale=0.5]{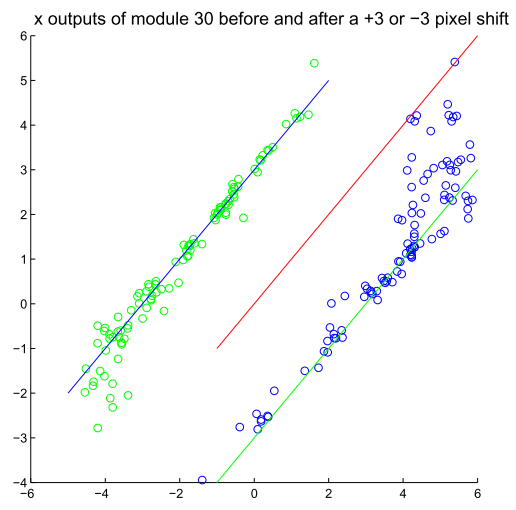}
    \caption{ (left) An illustration of three Transforming Auto-Encoder capsules, from \cite{Hinton2011TransformingAuto-encoders}.
    (right) A scatterplot showing, on the vertical axis, the x output of a capsule
    for an input image, and on the horizontal axis, the x output for the same capsule if the
    image is translated +3 (blue) or -3 (green) pixels in the x direction.}
    \label{fig:tae_example_1}
\end{figure}

For example, the Transforming Auto-Encoder (\textsc{TAE}) \cite{Hinton2011TransformingAuto-encoders} is
based on a series auto-encoder capsules. Each capsule is presented an image as input, and
is trained to reconstruct a translated version of that input image. The amount of horizontal
and vertical translation is fed as input into the network during training. An illustration
of three of this capsules is reproduced here on the left in Figure~\ref{fig:tae_example_1}. The units
in red represent \emph{recognition} units, and the units in green represent \emph{generation}
units. The variables $x$ and $y$ represent the translation of the input image. $\Delta x$ and
$\Delta y$ are fed as inputs to the network, and represent the $x$ and $y$ translation of the output
image. The $p$ variable represents the probability that the object represented by the capsule
is present in the image, and is multiplied with the output of the generation units to produce
the output of the capsule.

The authors demonstrate factor \textbf{distillation} on translated MNIST digits and on stereo 
image pairs of rotated cars. In each case, they find that the \textsc{TAE} learns a representation
of the translation or rotation that strongly correlates to the true value. This distillation
is demonstrated in the scatterplot of x outputs for a capsule in Figure~\ref{fig:tae_example_1} on the right.

\noindent \textbf{Similar Models.}
In \cite{Zhu2014Multi-ViewRepresentations}, a representation of faces is learned that separates
view from identity, and is trained to transform images in a similar way, using direct constraints 
on pose and equality constraints on identity. 
A similar approach is taken by, \cite{Yang2015Weakly-supervisedSynthesis}, which
generates a sequence of transformed views of an input image using a recurrent network.

\subsubsubsection{Disentangling Boltzmann Machines (Direct and Equality)}
The Disentangling Boltzmann Machine (\textsc{disBM}) \cite{Reed2014LearningInteraction} adds
supervisory constraints to a Restricted Boltzmann Machine with a multilinear bias. 
In the paper, the authors explore
various combinations of constraints, and explore the effect of constraining only
some of the factors in the representation.

\begin{figure}
    \centering
    \includegraphics[scale=0.45]{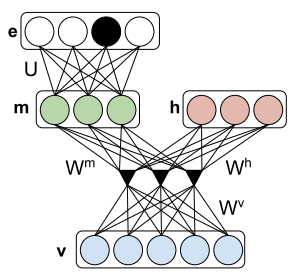}\qquad
    \includegraphics[scale=0.4]{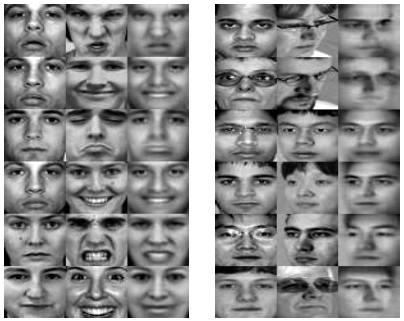}
    \caption{(left) A diagram of the \textsc{disBM} model architecture from \cite{Reed2014LearningInteraction}.
    (right) Example reconstructions where factors are mixed between observations.
    The left block of images represents transfer of expression, and the right block represents
    transfer of pose.
    In each row, the left hand image represents the identity. This identity is
    combined with the expression or pose of the middle image to form the image on the right.}
    \label{fig:disbm_example_1}
\end{figure}

Figure~\ref{fig:disbm_example_1} shows a diagram of the model architecture.
The hidden representation is divided into two sets of units, to model two factors.
In the diagram, \textbf{v} are the visible units, \textbf{h} are a set of hidden units
associated with one factor, and \textbf{m} are associated with another. 
\emph{Direct constraints} are applied by adding an extra layer on top of \textbf{m}
whose job is to predict the class label of that factor. This encourages the units
in \textbf{m} to represent information related to that class.
The version of the model with direct constraints is labeled \textsc{disBM (1)} in 
Figure~\ref{fig:model_chart}.

\emph{Equality constraints} are applied in two different ways. First, the authors
use a "clamping" procedure in which the units corresponding to a factor are encouraged
to be constant for the pair. For example, if \textbf{h} represents face identity, we
can clamp the \textbf{h} units for two images of the same individual. This is similar
in spirit to the clamping procedure we saw used in the IGN \cite{Kulkarni2015DeepNetwork}.

The second method for incorporating equality constraints involves adding a penalty term
to encourage the \textbf{h} representations to be more similar for the same individual. 
This is the same approach to the loss used by the \textsc{Siamese} network.

The version of the model with direct and equality constraints is labeled \textsc{disBM (1)} in 
Figure~\ref{fig:model_chart}, and the version with only equality constraints is \textsc{disbM (2)}.

The authors tested their models on two face datasets, the Toroto Face Dataset and Multi-PIE.
Both of these datasets have individuals with differing identities. For both datasets, \textbf{h}
units correspond to identity. For TFD, the \textbf{m} units correspond to expression, and for
Multi-PIE they correspond to pose.
On the right hand side of Figure~\ref{fig:disbm_example_1}, we can see an example of
pose and expression transfer between examples. The identity units and
pose or expression units are combined from two different images to form a new image that
demonstrates the disentangling of the factors.

\begin{figure}
    \centering
    \includegraphics[scale=0.35]{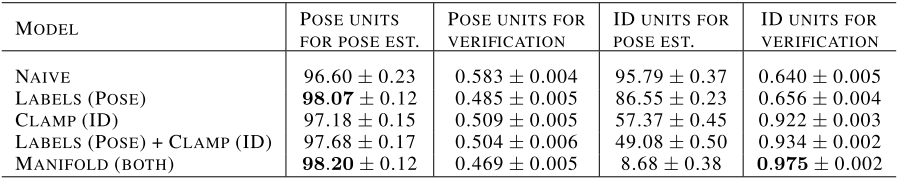}
    \includegraphics[scale=0.35]{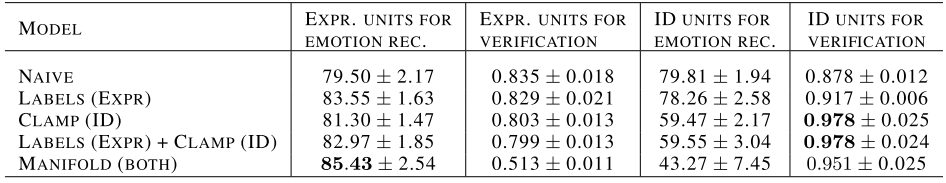}
    \caption{(top) Model results for the Multi-PIE dataset.
    (bottom) Model results for the Toronto Face dataset. } 
    \label{fig:disbm_example_2}
\end{figure}

The authors performed experiments with different levels of supervision, and investigated 
disentangling: the
effect of supervising one factor on disentangling of the other. 
They report performance on both prediction of the target factor value,
as well as prediction of the value of the \emph{non-target factor}. If the representation
is more factorial, then performance on the target factor should improve and performance
on the non-target factor should degrade.
The tables in Figure~\ref{fig:disbm_example_2} show predictive performance on each factor subspace,
for the Multi-PIE dataset on the top and the TFD dataset on the bottom.

In each table, the authors explore five different models. 
For example, for Multi-PIE, "Naive" refers to a mutlilinear
model without any supervision. "Labels (Pose)" is a model where the \emph{pose} variable
is supervised with direct constraints. "Clamp (ID)" uses the clamping procedure to learn
identity. "Labels (Pose) + Clamp (ID)" is a combination of those two, and 
"Manifold (Both)" uses the penalty term to learn both factors. The same naming
convention applies to TFD, but with "Pose" replaced with "Expression".

For each model and pool of units, the authors report test accuracy on a prediction task.
Test accuracy on pose and emotion prediction is a percentage correct, and accuracy
on verification is reported as a rate, from zero to one.

In the Multi-PIE table, we can see evidence of \textbf{distillation}: 
the supervised models generally have higher performance
than the "Naive" model. We also see that constraining a factor improves its performance ---
the pose units of "Labels (Pose)" predict pose better than the "Naive" model.
We see a similar effect with "Clamp (ID)". Also, as predicted, test accuracy on 
the prediction of the non-target factor goes down, indicating \textbf{disentangling}.

We can also see a \textbf{cross-over} effect:
the pose units of "Clamp (ID)" perform better at pose estimation than they do in
the "Naive" model. This effect is similar to what we observed in the \textsc{SSVAE} --- adding
supervised constraints to one factor tends to remove related variability 
from other factors.

The same pattern of results generally holds for TFD. We see supervised models
performing better than "Naive", and  we see decreasing performance on non-target tasks.
We can also see cross-over effects: the ID units of "Labels (Expr)" perform better
at verification than the ID Units of "Naive".
In three out of four cases, best test accuracy seems to come from manifold training.

\subsubsection{Desiderata for Supervised Models}
I have discussed various constraints, which have been incorporated into example models to
help discover factorial representations. However, the models vary along four
other dimensions as well. In order to completely characterize the models,
I summarize these dimensions of variation, since they are also relevant
for constructing factorial representations with supervisory signals.
Table~\ref{tab:supervised_chart} shows each model laid out on these dimensions.

\begin{table}[ht]
\centering
\caption{Listing of supervised models.}
{\rowcolors{2}{gray!10}{gray!0}
\begin{tabular}{|| m{28mm}|m{30mm}|m{16mm}|m{25mm}|m{15mm}|m{15mm}|m{16mm}||}
\hline
\textbf{Example}                                                                 & \centering \textbf{Constraint} & \textbf{Factor Agnostic} & \centering \textbf{Encoder}  & \centering \textbf{Prop. Dataset \newline Labeled} & \textbf{Prop. \newline Factors \newline Labeled} & \textbf{Generator} \\ \hline
\textsc{SSVAE} \cite{Kingma2014Semi-supervisedModels}                            & \hfil direct                   & \hfil                    & \hfil VAE                                      & \hfil \LEFTcircle                                  & \hfil \LEFTcircle                                & \hfil \checked                           \\
\textsc{disBM} (1) \cite{Reed2014LearningInteraction}                            & \hfil direct + equality        & \hfil                    & \hfil multilinear                              & \hfil \LEFTcircle                                  & \hfil \LEFTcircle                                & \hfil \checked                           \\
\textsc{DC-IGN} \cite{Kulkarni2015DeepNetwork}                                   & \hfil equality                 & \hfil                    & \hfil VAE                                      & \hfil \CIRCLE                                      & \hfil \CIRCLE                                    & \hfil \checked                           \\
\textsc{Siamese} \cite{Zeghidour2016JointNetworks}& \hfil equality                 & \hfil                    & \hfil neural net                               & \hfil \CIRCLE                                      & \hfil \CIRCLE                                    & \hfil                                    \\
\textsc{disBM} (2)                                                               & \hfil equality                 & \hfil                    & \hfil multilinear                              & \hfil \LEFTcircle                                  & \hfil \LEFTcircle                                & \hfil \checked                           \\
\textsc{TAE} \cite{Hinton2011TransformingAuto-encoders}                          & \hfil distance + equality      &                          & \hfil neural net                               & \hfil \CIRCLE                                      & \hfil \LEFTcircle                                & \hfil \checked                           \\
\textsc{mm t-SNE} \cite{VanDerMaaten2012VisualizingMaps}                         & \hfil distance                 & \hfil \checked           & \hfil lookup table                             & \hfil \CIRCLE                                      & \hfil \CIRCLE                                    &                                          \\
\textsc{Karaletsos} \cite{Karaletsos2015BayesianConstraints}                     & \hfil inequality               & \hfil                    & \hfil VAE                                      & \hfil \CIRCLE                                      & \hfil \LEFTcircle                                & \hfil \checked                           \\
\textsc{DVA} \cite{Reed2015DeepAnalogy-Making}                                   & \hfil analogy + equality       & \hfil                    & \hfil neural net                               & \hfil \CIRCLE                                      & \hfil \CIRCLE                                    & \hfil \checked                           \\ \hline

\end{tabular}
}
\label{tab:supervised_chart}
\end{table}

\subsubsubsection{Factor-Agnostic vs Factor-Specific}
Constraints can be \emph{factor-specific}, meaning each
label is directly tied to some known factor that is part of the representation.
Constraints can also be \emph{factor-agnostic}: the constraint has to
be satisfied by some factor, but the particular factor is not indicated by the label. For example,
people may attend to different features when rating bird image similarity: one person may attend
primarily to head color and another person might attend to tail color. In each case, the label collected
from the human is simply a similarity judgment, without any indication of which features the person was
attending to. In this case, the factors being compared are hidden, and the model has to infer which
factor a similarity constraint is associated with.
The only model fully described here that was factor-agnostic was \textsc{mm t-SNE}.
However, there are other models that use factor-agnostic constraints, such as 
\cite{Amid2015MultiviewMaps}.

Note that a factor-agnostic version of any
constraint multiplies the size of the constrained space by the number of factors in the 
representation.

\subsubsubsection{Encoder}
Each model has some kind of \emph{encoder} --- a means for mapping an observation
to its representation. 
Some models, such as \textsc{mm t-SNE}, simply have a fixed
lookup table. This avoids additional inductive bias but also disallows looking up a 
representation from an un-labeled observation.  This model does not make use
of direct features of the low-level data observations --- it uses only the
distance constraints. Therefore, without distance labels for an observation,
there is no representation for that observation.

The other models I discussed all have some kind of functional encoder. 
The various encoders in these models all
have different inductive biases: The \textsc{SSVAE}, \textsc{DC-IGN}, and 
\textsc{Karaletsos} models are
all based on the variational auto-encoder, which in turn uses the bias of Gaussian-
distributed factors. The \textsc{disBM} model is based on the multilinear architecture,
one of the combination biases described earlier.
The \textsc{Siamese} network, \textsc{DVA} and \textsc{TAE} are based on 
neural net architectures, and have lower unsupervised inductive bias for the encoder.

\subsubsubsection{Proportion of Dataset Labeled}
The strength of constraint is influenced by how much supervisory data we have.
If we have labels for each observation, we say the task is
\emph{fully-supervised}, and is marked with a \CIRCLE~in the table. 
A task where the dataset has labels for only a partial subset of the 
data is called \emph{semi-supervised}, and is marked with a \LEFTcircle.

\subsubsubsection{Proportion of Factors Labeled}
Stronger constraints also have labels for a larger portion of factors in the representation.
In most naturalistic environments, we cannot hope to label \emph{all} factors of variation --- for example,
consider the factors in the face and glasses dataset. It is easy to come up with labels for face identity
and glasses style. It is more difficult to label the position of the light source and the
face's pose, and there are likely more factors of variation beyond those which we might have trouble
even identifying. The more represented factors that are labeled, the stronger the constraints on the representation.
Models that make use of a full set of factor labels are marked with a 
\CIRCLE, and models that use a partial set of factor labels are marked with a \LEFTcircle. 

\subsubsubsection{Generator}
Some tasks, such as image generation, require a mapping
from a representation back to the observation.
Models suited to these tasks include a functional \emph{generator}; a function that maps the
representation back to the observation. 
All of the models except the \textsc{Siamese} network and \textsc{mm t-SNE} have 
a functional generator component.

A functional generator puts a heavy burden on the representation; it needs
to be able to reconstruct an input from its representation. 
In the case when only a subset of all factors are labeled, this means that
the other factors are unsupervised. In the case of \textsc{SSVAE} and \textsc{disBM}, 
we observed that cross-over effects of constraining one factor can help in learning 
the unsupervised factors.

A functional generator also makes producing a fully-labeled dataset very difficult.
It means that we need to know the full set of factors that contributed to generating
the observations. This explains why models that have both a generator and a fully
labeled dataset (\textsc{DC-IGN} and \textsc{DVA}) use synthetically-created
datasets.

\subsubsection{Hierarchy of Constraints} 
With so many different ways that supervision can vary, how can we rank strength
of supervision across different models? 
In this paper, I use the type of constraint (direct, equality, etc) as the primary 
indicator of supervision strength. 
Concerns related to the proportion of examples labeled in the dataset, or the proportion
of factors labeled, are fairly dataset-specific, but I am interested in the differences
between constraints, not the particular test environments of the papers.
Whether the constraint is factor-agnostic or not is important. However, as noted earlier,
this concern will not move a constraint type up or down in the overall ranking, but will
make any given constraint type slightly weaker.
A model can also include supervisory information using multiple constraint types.
When constraint types are mixed, the ranking of a model in the chart is determined 
by the weakest type of constraint used.


\section{Discussion}  

\subsection{Summary of Biases}
A wide variety of biases can help discover factorial representations,
and each bias comes with an associated set of generative assumptions.
I demonstrated how sparsity bias in \textsc{ICA} can help to learn
factorial representations when the generative environment is sparse.
Likewise, an invariance bias helps bring coherency in factors that can
be expressed in a variety of ways.
Different combination biases are also shown to lead
to factorial representations when the environment is matched to the generative
assumptions. 

Supervisory bias is used to shape representations to match prior knowledge
about the generative environment. Different types of supervised constraints
are associated with different types of prior knowledge, and vary in how
strongly the constrain the representation.

The biases described here can be combined together in various ways.
For example, sparse distribution bias
is combined with invariance bias in the \textsc{ISA} model,
with multilinear bias in a sparse bilinear model, and with 
hierarchical layers in \textsc{R-ICA}.
Likewise, supervised constraints can be combined with an encoder 
with distributional, invariance, or combination biases.
For example, \textsc{DC-IGN} combines the bias of Gaussian-distributed, independent, factors
with supervised equality constraints. \textsc{disBM} combines multilinear
factor combination bias with equality and direct constraints.
In general, models with supervised bias make use of low unsupervised 
bias --- with enough supervised constraints, super-strong unsupervised
bias is no longer necessary.

The key to discovering factorial representations is to match unsupervised
inductive bias to the environment, and, if the unsupervised sources 
of bias are not sufficient, use additional supervisory signals to learn.

\subsection{Effect of Bias on Factorization}
Since there are numerous examples of different combinations of biases, the
natural question is: how do they compare with each other? Which bias leads
to representations where factors are most distilled and disentangled from one
another?
Can we lay out a space where different levels
of supervised constraints correspond with degrees of factorization?

Unfortunately, the models cannot be compared to one another directly given the
evidence from the papers. They make different assumptions about the environment, 
work in different data domains (e.g. image, audio), and use different data sets. 

However, there is some indication that stronger constraints lead to 
more cross-over effects, which indicates better factorization.
Each model demonstrates some evidence of factorization in the representation,
and most models show evidence for distillation and disentangling.
Evidence of a cross-over benefit is found for the \textsc{SSVAE}, \textsc{disBM},
and, to a lesser extent, \textsc{Siamese} models. Their positions in the model chart
in Figure~\ref{fig:model_chart}, indicate that these models all have fairly strong 
supervised constraints, and \textsc{disBM} has a strong combination bias.
While it is possible that the other papers could have shown cross-over
effects but did not report them, it seems plausible that higher levels of 
bias should cause stronger factorization in the representation.

Another hint that stronger constraints lead to better factorization is
in the datasets used for training in the papers. The only papers that used
partially labeled datasets were \textsc{SSVAE} and \textsc{disBM}, which
also used the strongest biases.

The model with the least evidence for factorization was \textsc{mm t-SNE}, despite
the use of the fairly strong distance constraints. This can be attributed to the use
of a dataset with a very high number of factors (in this case, topics), factor-agnostic
constraints, and a very large space of inputs (thousands of words).

\subsection{Directions for Future Research}

\subsubsection{Factorial Representations in Non-Factorial Environments}
None of the papers considered the effect of data set sampling bias
or non-factorial environments. Is a supervisory signal always necessary
to learn factorial representations, or are some unsupervised inductive biases
sufficient?
An interesting future project could investigate the effect of the level of
factorization of a dataset on a model's ability discover a factorial representation.
For example, suppose we could vary the amount of correlation between identity and presence of glasses
in a face dataset, and use a variable number of supervised constraint examples to learn a factorial
representation. How many supervised examples are required to learn in the case when there is
no correlation, in the case of some correlation, or in the case where 
identity and glasses are completely correlated?
Does this relationship change with constraint type? Does this relationship change
with our choice of unsupervised inductive bias?

\subsubsection{New Combinations of Biases}
The model chart in Figure~\ref{fig:model_chart} hints at some interesting
unexplored combinations of biases. 

Several of the unsupervised biases could be combined with an autoencoder to
further reduce factor combination bias and help with invariance.
For example, we could combine the bias
of multilinear combination with an autoencoder architecture. Multilinear
models are pretty wasteful in terms of model size --- they need to learn one
component vector for every combination of latent factors. An autoencoder could
be both more efficient in representing these combinations, and have a lower bias
in terms of how the combinations are combined.
Sparse autoencoders combine the sparsity bias with an autoencoder. We could
similarly extend an autoencoder to use pooled/block sparsity of \textsc{ISA}
to find multi-dimensional sparse factors.
We could even experiment with merging different combination biases, e.g. 
use the non-negativity bias of \textsc{NMF} in a \textsc{Multilinear} model
to create a functional parts model where the parts can interact multiplicatively.

We could similarly create new combinations of supervised biases. For example,
we could apply the "learning to transform" model of \textsc{TAE} to direct constraints,
and learn a model that represents the presence of glasses by transforming an image
of a person with glasses to an image of the same person without.

Relatively few papers use factor-agnostic constraints, despite their potential usefulness.
It is difficult to identify all the dimensions humans might use to compare examples.
None of the papers that use factor-agnostic constraints learn a representation 
with a functional encoder.
It would therefore be interesting to extend a model like \textsc{Karaletsos} to handle
factor-agnostic triplets.



\bibliographystyle{plain}
\bibliography{references,mendeley}
\end{document}